\documentclass[lettersize,journal]{IEEEtran}
\usepackage{soul} 
\usepackage[flushleft]{threeparttable}
\usepackage{arydshln}
\usepackage{color, xcolor} 
\usepackage{amsmath,amsfonts}

\usepackage{colortbl}
\definecolor{spink}{rgb}{0.99, 0.91, 0.95}
\definecolor{sblue}{rgb}{0.94, 0.97, 1.0}
\definecolor{syellow}{rgb}{1.0, 0.55, 0.0}
\usepackage{xspace}
\usepackage{array}
\usepackage{multicol}

\usepackage{stfloats}
\usepackage{url}
\usepackage{verbatim}
\usepackage{cite}
\usepackage{booktabs}
\usepackage{amsmath}
\usepackage[ruled,linesnumbered]{algorithm2e}
\usepackage{sidecap}
\usepackage{multirow}
\usepackage{floatrow}
\usepackage{wrapfig}
\usepackage{amssymb}
\usepackage{graphicx}
\usepackage{lipsum}
\usepackage{tikz}
\usepackage{pgfplots}
\usetikzlibrary{patterns}
\pgfplotsset{compat=1.8}
\pgfmathdeclarefunction{fpumod}{2}{%
    \pgfmathfloatdivide{#1}{#2}%
    \pgfmathfloatint{\pgfmathresult}%
    \pgfmathfloatmultiply{\pgfmathresult}{#2}%
    \pgfmathfloatsubtract{#1}{\pgfmathresult}%
    \pgfmathfloatifapproxequalrel{\pgfmathresult}{#2}{\def\pgfmathresult{5}}{}%
}
\pgfplotsset{boxplot legend/.style={
    legend image code/.code={
        \draw[#1] (0cm,0cm) rectangle (0.6cm,0.3cm)
        (0.3cm,0cm) -- (0.3cm,-0.1cm) (0.1cm,-0.1cm) -- (0.5cm,-0.1cm)
        (0.3cm,0.3cm) -- (0.3cm,0.4cm) (0.1cm,0.4cm) -- (0.5cm,0.4cm);
    },
}}
 
\usepgfplotslibrary{statistics}
\usetikzlibrary{pgfplots.statistics}
\usepackage{subfigure}
\usepackage{algorithmicx}
\usepackage{algcompatible}
\usepackage{algpseudocode}
\usepackage{lineno,hyperref}
\newcommand{\msm}[1]{{\color{black} #1}} 
\pgfplotsset{width=7cm,compat=1.8}
\pgfplotsset{%
  colormap={whitered}{color(0cm)=(white);
  color(1cm)=(blue!75!red)}
}
\newcommand{\mm}[1]{{\color{black} #1}} 
 \usepackage{pgfplotstable} 

\ifCLASSINFOpdf

\else
 
\fi

\usepackage{etoolbox}
\makeatletter
\patchcmd{\@makecaption}
  {\scshape}
  {}
  {}
  {}
\makeatother

\begin{document}
\title{3D Harmonic Loss: Towards Task-consistent and Time-friendly 3D Object Detection on Edge for V2X Orchestration}


\author{Haolin Zhang, M S Mekala, Zulkar Nain, Dongfang Yang, Ju H. Park,
Ho-Youl Jung
\thanks{Note: Haolin Zhang, M S Mekala are both contributed equally for accomplishing the targets. Haolin Zhang is with Institute of Artificial Intelligence and Robotics, Xi'an Jiaotong University, China.
(E-mail:zhanghaolin@xjtu.edu.cn)}
\thanks{M S Mekala, and Ho-Youl Jung are with department of Information and Communication Engineering, Yeungnam University, Gyeongsan 38544, Korea as well as RLRC for Autonomous Vehicle Parts and Materials Innovation, Yeungnam University, Gyeongsan 38544, Korea (E-mail: msmekala@yu.ac.kr \& hoyoul@yu.ac.kr (corresponding author)).}
\thanks{Dongfang Yang, Chongqing Chang'an Automobile Co., Ltd. Chongqing, China. (E-mail: yangdf@changan.com.cn).}
\thanks{Ju H. Park,
Department of Electrical Engineering, Yeungnam University, Gyeongsan 38544, Korea.  (E-mail: jessie@ynu.ac.kr).}
%
}

\maketitle

\maketitle

\begin{abstract}
\mm{Edge computing-based 3D perception has received attention in intelligent transportation systems (ITS) because real-time monitoring of traffic candidates potentially strengthens Vehicle-to-Everything (V2X) orchestration. Thanks to the capability of precisely measuring the depth information on surroundings from LiDAR, the increasing studies focus on lidar-based 3D detection, which significantly promotes the development of 3D perception. Few methods met the real-time requirement of edge deployment because of high computation-intensive operations. Moreover, 
an inconsistency problem of object detection remains uncovered in the pointcloud domain due to large sparsity. This paper thoroughly analyses this problem, comprehensively roused by recent works on determining inconsistency problems in the image specialisation. Therefore, we proposed a \textit{3D harmonic loss} function to relieve the pointcloud based inconsistent predictions.
Moreover, the feasibility of \textit{3D harmonic loss} is demonstrated from a mathematical optimization perspective. The KITTI dataset and DAIR-V2X-I dataset are used for simulations, and our proposed method considerably improves the performance than benchmark models. Further, the simulative deployment on an edge device (Jetson Xavier TX) validates our proposed model's efficiency.}


\end{abstract}
\IEEEpeerreviewmaketitle

\mm{
\section{Introduction}
\IEEEPARstart{B}{ackground}: Edge computing-based computer vision technology has received global attention for strengthening V2X orchestration and autonomous driving systems (ADS). V2X and ADS are multi-disciplinary research domains where the vehicles and infrastructures collect and analyze the data \cite{g3} to make the vehicle move through an intelligent decision-making system. The decision-making system fed the captured data (surrounding areas, including road structure and traffic information). The data is analyzed to make necessary decisions for the vehicle move based on cloned effective object detection methods through Road Side Units (RSU) and On Board Units (OBU) to identify and localize the traffic candidates.

\textbf{\textit{Motivation:}} Let us consider a vehicle movement based on event-trigger analysis using traffic information. In this process, the traffic data collects through surveillance devices or on-vehicle sensors such as LiDAR and cameras. The important lidar data (pointcloud) is being analyzed by edge devices to accomplish the target with low latency. However, the computation-intensive services are being offloaded to the server to meet the application deadline. A lidar 3D object detection technology has become prominent because of its low price, increased perception of distant objects, and robust characteristics. For pinpoint communication from vehicle to vehicle (V2V) and vehicle to infrastructure (V2I), the recognized and localized vehicle has become an essential factor in measuring the surroundings and infrastructure through RSU-LiDAR deployment. Thus, developing and deploying an efficient and robust 3D detector based on lidar data is an essential and worthy research direction for strengthening V2X.

\textbf{\textit{Problem of task inconsistency and time delay:}} Modern object detection tasks have diverged into several sub-tasks (e.g. object localization, classification, direction estimation, etc.). In the 2D image domain, most 2D detection considers sub-tasks (classification and localization) independently, resulting in the output of inconsistently unexpected predictions with high classification confidence but insufficient localization after post-processing (e.g. Non-Maximum Suppression), as shown in Fig.\ref{fig:fig1}(a). Such inconsistency problem in 2D object detection has recently been fully discussed and partially solved in \cite{gfl,gflv2,harmony,paa}. Turning to the pointcloud domain, the guesswork and similar inconsistency issues remain effects the 3D detection accuracy, as shown in Fig.\ref{fig:fig1}(b). Real-data experiments further confirmed our conjecture in Fig.\ref{fig:fig1}(c). \msm{Most recent lidar-based 3D object detection methods \cite{second,pointgnn,pointrcnn,3dssd,voxelnet,voxelrcnn,part2,pvrcnn,voxel-set,sessd,density-voxel,centerpoint} concentrated on achieving the best mAP and treated it as a phenominal benchmark for representing model accuracy. Nevertheless, time consumption, quality of experience (QoE), and service reliability factors are essential to showcase the model performance, which has not been addressed. Especially for real-time applications, as in V2X, there is a scope to design a cost-effective, task-friendly and task-consistent detection solution with fast run-time and low error rate. On the one side, several researchers \cite{ciassd,cl3d} noticed the problem of task inconsistency. Nevertheless, their solutions relied on 
additional modules with extra inference time-cost, contrary to our application target of pursuing less computational burden. On the other side, some recent works\cite{pp,pixor,bevdet,fast,rt3d,ppdeploy} attempted to improve the performance of deployment metrics such as computational burden and execution latency. However, these methods have yet to achieve an adequate trade-off between detection accuracy and time consumption towards edge device-based simulations for real-time applications.}

\textbf{\textit{Our solutions:}} We derived solutions from the learning optimization perspective to solve the above drawbacks for better edge-computing object detection performance. Firstly, by drawing the lessons from the inconsistent prediction problem in camera-based 2D detection, we indicate a similar inconsistency problem in lidar-based 3D detection. This problem gradually leads to the inaccuracy of the prediction in actual applications and is worth being discovered and resolving. To alleviate inconsistent predictions of 3D detectors, we analyze the cause of the inconsistency problem through the respective characteristics of the image and point cloud. Inspired by the solution in image domain \cite{harmony},  we extend the 2D solution to 3D detection and propose \textit{3D harmonic loss}, a task-consistent learning strategy for optimizing pointcloud-based 3D detectors. It is worth mentioning that our solution, \textit{3D harmonic loss}, not like previous solutions\cite{ciassd,cl3d}, only works for model training and does not bring any extra time-cost to model inference. Secondly, a thorough mathematical analysis is conducted to explain and demonstrate the effectiveness of \textit{3D harmonic loss}. Experiments on KITTI 3D/BEV detection dataset\cite{kt} further validate that the proposed strategy can achieve a noticeable performance improvement. Third, our proposed model is deployed on the edge device (Jetson Xavier TX) for simulation, and it achieves an ideal trade-off between time efficiency and detection accuracy. 
We deploy the proposed detector on edge devices (Jetson Xavier TX) for realistic simulations to meet the lightweight design and edge-computing benchmark metrics.

	\begin{figure*}[t]
		\centering
		\includegraphics[width=1.0\linewidth]{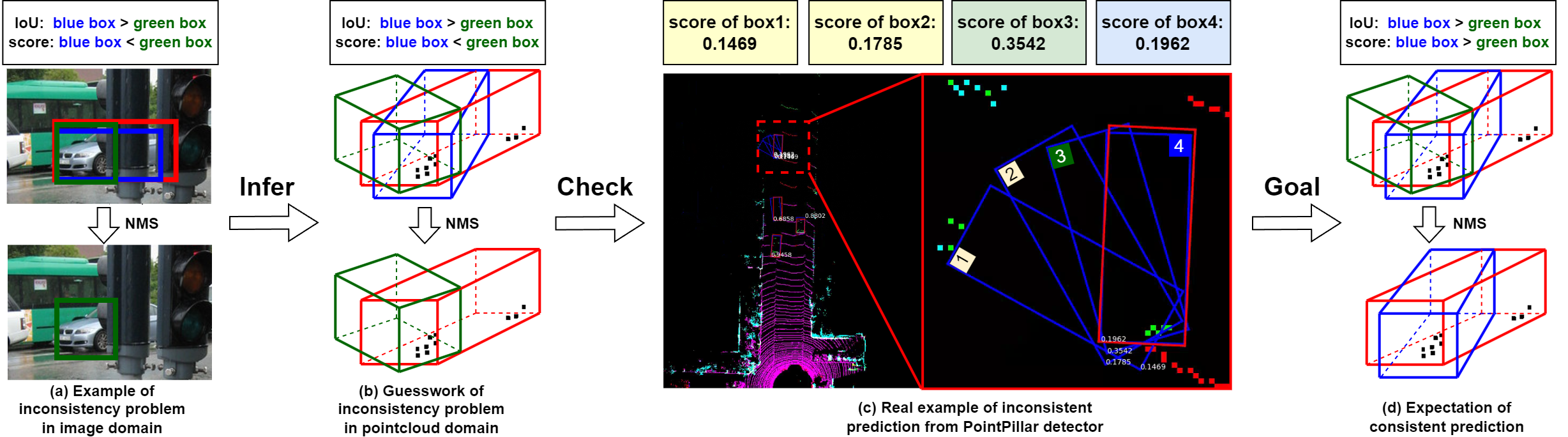}
		\caption{Illustration of the inconsistency problem in object detection. (a) Example of inconsistency problem in image domain: inconsistent bounding boxes with high classification score but low IoU (compared to groundtruth (\textcolor{red}{red box})) in 2D detection, which leads to the suboptimal output (\textcolor{green}{green box}) after post processing (NMS). (b) Guesswork of the similar inconsistency problem in pointcloud-based 3D detection.  (c) Real example of inconsistency problem from the PointPillar\cite{pp}. (d) Expectation of consistent prediction: a better 3D detector is expected to harmonize the localization and classification of predicted objects, resulting in the reasonable output (\textcolor{blue}{blue box}). Our work focuses on how to alleviate the inconsistent predictions in pointcloud domain, to achieve the expected predictions in real-world applications.}
		\label{fig:fig1}
	\end{figure*}

Our contributions are as follows. 
\begin{enumerate}
\item Develop a \textit{3D harmonic loss} method for alleviating inconsistent predictions inspired by related ideas from 2D detection. Thus, we level up the 2D solution to lidar-based 3D detection to map both two-stage and one-stage 3D detection models' learning accuracy without extra time-cost on inference.
\item Experiments on KITTI Dataset \cite{kt} and DAIR-V2X-I Dataset \cite{dair} demonstrate our proposed work's effectiveness for both on-vehicle and on-infrastructure object detection. Especially for industrial-popular lidar-based detectors such as SECOND \cite{second}, and PointPillar \cite{pp} are considered to showcase the significant margin of mean average precision (mAP) improvement concerning the proposed \textit{3D harmonic loss}. 

\item Realistic simulations by deploying our proposed lightweight detector on the Jetson Xavier device further verify and realise that our solutions are time-friendly and task-consistent towards 3D detection for real applications.

\end{enumerate}
The paper continues as Section \ref{sec2} that briefs the extant approaches research gaps. Section \ref{sec3} represents the proposed work in detail. Section \ref{sec4} represents the proposed method's effectiveness using qualitative and quantitative analysis. Section \ref{sec5} concludes the manuscript.

\section{Related Work}\label{sec2}

\subsection{LiDAR-based 3D object detection}


3D object detection has become more popular due to pointcloud-based deep learning models through various frameworks. Usually, there are two types of frameworks (one-stage and two-stage). One-stage methods predict the 3D bounding boxes (bboxes) of objects instantly. Some of the methods are points-based; such as 3DSSD \cite{3dssd} is designed based on PointNet \cite{pointnet} architecture, and PointGNN network \cite{pointgnn} is developed based on graph neural network. Where raw lidar pointcloud is fed to the learning model for 3D shape predictions. Some other voxel-based methods, such as VoxelNet \cite{voxelnet}, first rasterizes the lidar pointcloud into 3D voxels to decrease the input memory occupation. After that, voxel features are fed to a region proposal network with 3D convolutions for 3D detection. For instance, SECOND \cite{second} is a improved time efficiency approach which is based on VoxelNet \cite{voxelnet} by proposing sparse 3D convolutions. Nevertheless, the time performance of one-stage 3D detection is not desirable enough. Besides, VoTr \cite{trs} and VoxSeT \cite{voxel-set} not only adopt a voxel-based one-stage method but also introduce transformer \cite{attention} architecture for better accuracy. However, heavy parameters and complicated operations significantly impair the time performance of 3D detection. Compared to the methods above, PointPillar \cite{pp} transforms 3D pointcloud to 2D voxels, followed by highly efficient 2D convolutions to achieve real-time performance and easy deployment of 3D detection \cite{ppdeploy}. 
By contrast, two-stage detectors \cite{pointrcnn,voxelrcnn,part2,pvrcnn,centerpoint} predict the Region-of-Interests (ROIs) in the first stage; the refinement network server precisely detects the objects based on ROIs in the second stage. However, most two-stage methods cannot satisfy the speed needed in real-time applications. Although several methods, e.g. CenterPoint \cite{centerpoint}, has the advantage of fast feature encoding and the light refinement head in their network design. However, the research gap (time-cost comparison) remains similar to some one-stage detectors such as PointPillar \cite{pp}.

\msm{The fusion of image and pointcloud \cite{fpointnet,ff,clocs,IF,ppint} is an adequate mechanism to achieve better surrounding perception and accuracy for 3D detection. However, the fusion methods are more complex and consume abnormal time in real-time applications than pure lidar-based detection. Therefore, the fusion methods are not mainly discussed but we considered some fusion methods in our experiment to showcase the accuracy accomplishments.}

\subsection{Inconsistency problem in object detection}
The inconsistency problem was first discovered in 2D object detection methods of the image domain, as shown in Fig.\ref{fig:fig1}(a). Early 2D object detection methods independently treat object classification and object localization (two sub-tasks of 2D object detection) in the model training phase, which causes inconsistent predictions in the inference phase. Recent works \cite{gfl,gflv2,harmony,paa} focused on this problem in the image domain and tried to bridge the gap between different sub-tasks of 2D detection. In \cite{gfl,gflv2}, a Generalized Focal Loss method is developed, which is not reached the targeted accuracy, and an improved version of the Focal Loss method is developed in \cite{focal-loss} to guarantee coherent 2D detection. \msm{In \cite{paa}, a PAA method is introduced with an additional module to predict IoU for the rational selection of positive training samples. The inconsistency problem in 3D pointcloud object detection mechanisms causes low object detection reliability and quality of experience. For instance, several works \cite{ciassd,cl3d} attempted to solve it. However, their solutions, similar to PAA \cite{paa}, consume extra time-cost for a supplementary branch of predicting IoU and additional operations in post-processing, which are not applicable for real-time environments. Our proposed solution deals with the inconsistency problem well in the pointcloud domain without introducing any additional burden for model inference and deployment.}






\section{Proposed Work}\label{sec3}

In this section, the proposed method 3D harmonic loss is formulated in both theoretical and mathematical optimization perspective.


Fig. \ref{fig:fig1}(d) shows our goal to achieve consistent predictions in 3D detection. From revisiting the learning loss function (Eq\ref{sota}) for a positive training sample $i$, in many extant methods \cite{pp,pointrcnn,part2,second}, the cause of inconsistency problem is excavated: three sub-tasks (classification, localization (regression) and direction estimation) of 3D object detection are treated and supervised independently. 

	\begin{figure*}[!t]
		\centering
		\includegraphics[width=1.0\linewidth]{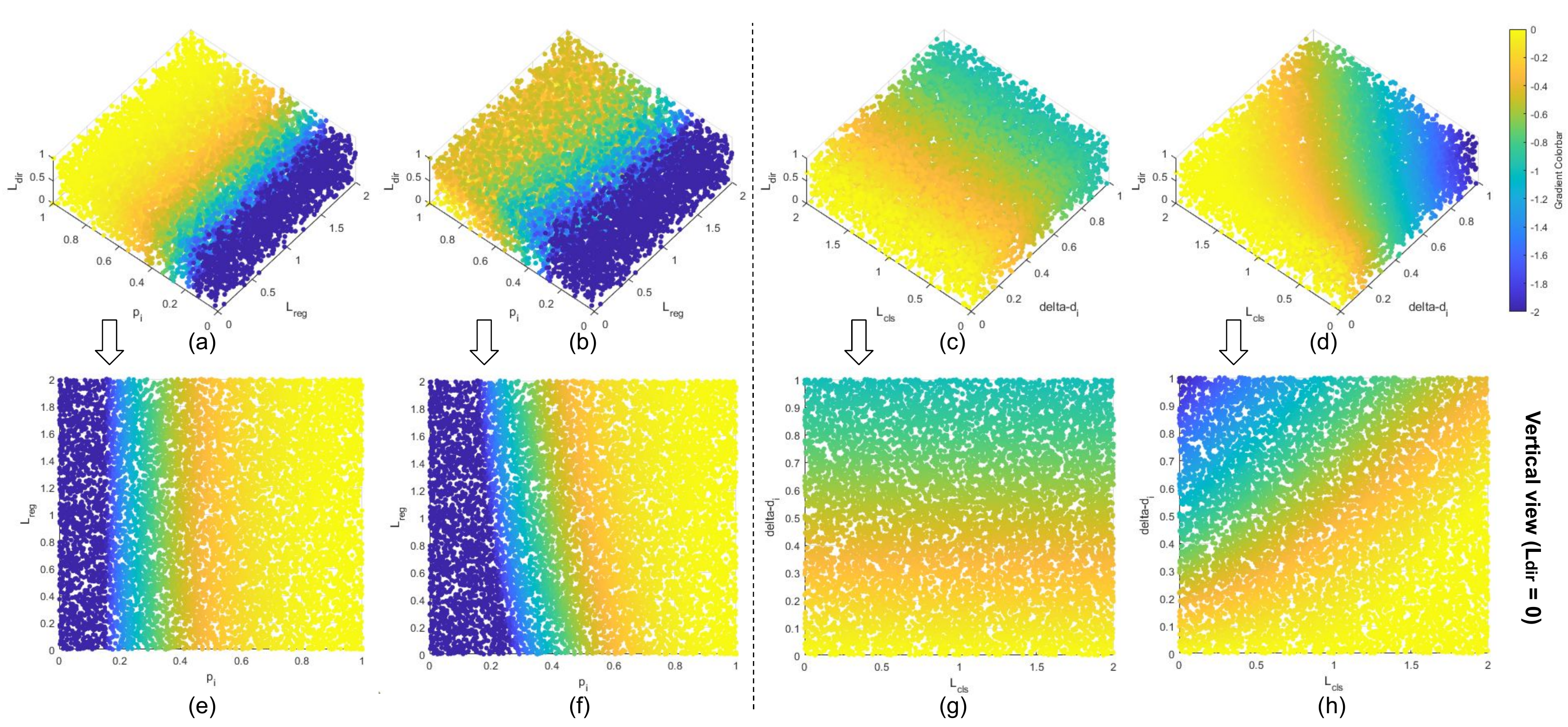}
		\caption{Visualization of the gradients from 3D detection loss related to different sub-tasks (object classification and object localization (regression)). (a) is drawn with gradients from classification part in common 3d detection loss. (b) is drawn with gradients from classification part in our proposed 3d harmonic loss. (c) is drawn with gradients from regression part in common 3d detection loss. (d) is drawn with gradients from regression part in our proposed 3d harmonic loss. For better view, (e), (f), (g) and (h) show their vertical forms. The color intensity indicates the value of gradients (see colorbar).}
		\label{fig:fig3}
	\end{figure*}
\msm{
\begin{equation}
   \begin{array}{l}\label{sota}
L_{3D}^{i} = L_{cls}\left(p_{i},p_{i}^{gt}\right) + L_{reg}\left(d_{i}^{'},d_{i}^{gt} \right)\,\, +  L_{dir}\left(p_{i}^{'},p_i^{{'}^{gt}} \right)
\end{array}
\end{equation}

Where $p_i$ is softmax classification score, $p_i^{'}$ is softmax direction score. Also $p_{i}^{gt}$ and $p_i^{{'}^{gt}}$ are the ground truths for classification and direction estimation respectively. Consequently, the classification loss ($L_{cls}(p_i,p_i^{gt})$) for positive training samples ($p_{i}^{gt}$=1)  uses focal loss\cite{focalloss}, which is derived as follows 
\begin{equation}
    {L_{cls}}\left( {{p_i}, p_i^{gt}} \right) =  - \alpha {\left( {1 - {p_i}} \right)^\gamma }\log \left( {{p_i}} \right)
\end{equation}
In continuation, the regression loss ${L_{reg}}$ uses $\text{Smooth}L_1$\cite{fasterrcnn} as follows 
\begin{equation}
{L_{reg}}\left( {{d_i^{'}},d_i^{gt}} \right) = \sum\limits_{{d_i^{'}} \in \left( {{x_i^{'}},{y_i^{'}},{z_i^{'}},{l_i^{'}},{w_i^{'}},{h_i^{'}},{\theta _i^{'}}} \right)} {\text{Smooth}{L_1}\left( {\Delta_{d_i}} \right)}  \end{equation}

\begin{equation}
    \text{Smooth}{L_{1}}\left( \Delta_{d_i} \right)= \left\{\begin{matrix}
0.5\Delta_{d_i}  & if\left | \Delta_{d_i}  \right | <1& \\ 
\left | \Delta_{d_i}  \right |-0.5& \text{others} & 
\end{matrix}\right.
\end{equation}
}


Where $\Delta_{d_i}$ is the difference between the set of attributes (${x_i^{'}},{y_i^{'}},{z_i^{'}},{l_i^{'}},{w_i^{'}},{h_i^{'}},{\theta _i^{'}}$) of predicted offsets ${d_i^{'}}$ and ground truth offsets $d_i^{gt}$}, which is determined by the parameters (${X_i^{gt}},{Y_i^{gt}},{Z_i^{gt}},{L_i^{gt}},{W_i^{gt}},{H_i^{gt}},{\alpha _i^{gt}}$) of ground truth boxes and the parameters (${X_i},{Y_i},{Z_i},{L_i},{W_i},{H_i},{\alpha _i}$) of anchor boxes as follows
\begin{equation}
\begin{array}{l}
{x_i}^{gt} = \frac{{X_i^{gt} - {X_i}}}{{\sqrt {{{\left( {{W_i}} \right)}^2} + {{\left( {{L_i}} \right)}^2}} }}\,,\,\, {y_i}^{gt} = \frac{{Y_i^{gt} - {Y_i}}}{{\sqrt {{{\left( {{W_i}} \right)}^2} + {{\left( {{L_i}} \right)}^2}} }}\,\,\,\,\\
{z_i}^{gt} = \frac{{Z_i^{gt} - {Z_i}}}{{{H_i}}},\,\,\,\,{w_i}^{gt} = \frac{{W_i^{gt}}}{{{W_i}}}, {l_i}^{gt} = \log \frac{{L_i^{gt}}}{{{L_i}}}\,\,\,\\
\,{h_i}^{gt} = \log \frac{{H_i^{gt}}}{{{H_i}}}, {\theta _i}^{gt} = \sin \left( {\alpha _i^{gt} - {\alpha _i}} \right)\,
\end{array}    
\end{equation}
\mm{Inconsistent treatment towards the different sub-tasks leads to inconsistent inference results. Previous work \cite{harmony} provides a good prior attempt. However, it only focused on 2D detection with the image source and only considered generalizing basic loss functions (e.g. cross-entropy loss, L1 loss, and IoU loss, etc.). For lidar-based 3D detection, with different data modalities (pointcloud) and more dimensions of prediction (direction estimation, height and depth, etc.), we proposed 3D harmonic loss (see \textbf{Theorem}). Three dynamic factors $1+\beta _r$, $1+\beta _c$ and $1-{\frac{{{\beta _r} + {\beta _c}}}{{{\beta _{dir}}}}}$ are used to make certain that the model learning is conducted in a way of consistency, where $1+\beta _r$ and $1+\beta _c$ work together to ensure \textbf{mutual-consistency}, and $1-{\frac{{{\beta _r} + {\beta _c}}}{{{\beta _{dir}}}}}$ guarantees \textbf{intrinsic-consistency}. \msm{Mutual consistency ensures when classification optimisation is inadequate, and the factor derived from the classification part will supervise the regression part and vice versa. Intrinsic consistency guarantees that the direction estimation part is always supervised by both classification and regression parts. For example, a perfect direction estimation should be consistent with clear class recognition and unambiguous boundary regression.} 
Moreover, our proposed learning mechanism can be well adapted to focal loss (object classification), $\text{Smooth}L_1$ (object localization) and binary cross-entropy (object direction estimation), and these three loss functions are frequently used in 3D detection training. In a word, our method considers all sub-tasks in a harmonious manner.}
\\


\mm{
\textbf{Theorem:} 3D harmonic loss $$\begin{array}{l}
L_{3D - Har}^i = \left( {1 + {\beta _r}} \right) \times {L_{cls}}\left( {{p_i},p_i^{gt}} \right) + \left( {1 + {\beta _c}} \right)\\
\,\,\,\,\,\,\,\,\,\,\,\,\,\,\,\,\,\, \times L_{reg}\left(d_{i}^{'},d_i^{gt} \right)\,\, + \left(1 - \frac{\beta_{r} + \beta _{c}}{\beta_{dir}} \right) \times L_{dir}\left(p_{i}^{'},\,p_i^{{'}^{gt}} \right)
\end{array}$$ 

Where 
\begin{equation}
    {\beta _r} = {e^{ - {L_{reg}}\left( {{d_i^{'}}, d_i^{gt}} \right)}},{\beta _c} = {e^{ - {L_{cls}}\left( {{p_i}, p_i^{gt}} \right)}}
\end{equation}
We consider $\beta_{dir}=2$ in our 3D harmonic loss simulation. When the training sample is positive, it is consistently supervised by classification loss ${L_{cls}}\left( {{p_i}, p_i^{gt}} \right)$, regression loss ${L_{reg}}\left( {{d_i^{'}},d_i^{gt}} \right)$, \msm{and direction loss $L_{dir}\left(p_{i}^{'},p_{i}^{'gt} \right)$.} The following proofs are mathematically derived and explained briefly to represent the effectiveness of 3D harmonic loss.

\textbf{Proof-1:}
Let us assume, the $i^{th}$ training sample is positive, then $p_i^{gt}=1$, and $\alpha=0.25$, $\gamma=2$ are for the focal loss to analyse the effectiveness of 3D hormonic loss towards classification loss which is derived as follows
\begin{equation}\label{cls}
    \begin{array}{l}
\frac{{\partial {L_{cls}}\left( {{p_i},p_i^{gt}} \right)}}{{\partial {p_i}}} = \frac{{\partial \left[ { - \alpha p_i^{gt}{{\left( {1 - {p_i}} \right)}^r}\log \left( {{p_i}} \right)} \right]}}{{\partial {p_i}}}\\
\,\,\,\,\,\,\,\,\,\,\,\,\,\,\,\,\,\,\,\,\,\,\,\,\,\,\,\,\,\,\,\,\,\, =  - \frac{{\left( {1 - {p_i}} \right)\left( {\frac{1}{{{p_i}}} - 1 - 2\log \left( {{p_i}} \right)} \right)}}{4}\mathop  = \limits^{suppose} J\left( {{p_i}} \right)
\end{array}
\end{equation}
\begin{equation}\label{cls0}
    \begin{array}{l}
{\beta _c} = {e^{ - {L_{cls}}\left( {{p_i},p_i^{gt}} \right)}} = {e^{ - \frac{1}{4}{{\left( {1 - {p_i}} \right)}^2}\log \left( {{p_i}} \right)}} \mathop =\limits^{suppose} K\left( {{p_i}} \right)
\end{array}
\end{equation}
with the point derivation
\begin{equation}\label{cls1}
    \frac{{\partial {\beta _c}}}{{\partial {p_i}}} = -\frac{{{(1-p_i)^2}}}{4}{p_i}^{\frac{{-{(1-p_i)^2}}}{4}-1}  =  \nabla K\left( {{p_i}} \right)\end{equation}
Based on Eq.\ref{cls}, Eq.\ref{cls0} and \ref{cls1} the gradient backpropagation from the classification part is represented as Eq.\ref{bpcls}.
\begin{equation}
    \begin{array}{l}\label{bpcls}
\frac{{\partial H_{3D-Har}^i}}{{\partial {p_i}}} = \left( {1 + {e^{ - {L_{reg}}}}} \right)J\left( {{p_i}} \right) + \\
{L_{reg}}.\nabla K\left( {{p_i}} \right) - {L_{dir}}\left( {{p_i^{'}},p_i^{{'}^{gt}}} \right).\frac{{K\left( {{p_i}} \right)}}{{{\beta _{dir}}}}
\end{array}
\end{equation}
Note that, in our experiment $\beta _{dir}=2$, the gradient backpropagation from the classification result is highly associated to $$\begin{array}{*{20}{c}}
{{L_{reg}}}\\
 \uparrow \\
{regression}
\end{array},\begin{array}{*{20}{c}}
{{p_i}}\\
 \uparrow \\
{classification}
\end{array},\begin{array}{*{20}{c}}
{{L_{dir}}}\\
 \uparrow \\
{direction\ estimation}
\end{array}$$

\textbf{Analysis-1: }
Consequently, the Eq.\ref{bpcls} outcomes illustrated in Fig.\ref{fig:fig3}(b) with averagely sampling ten thousands data. The color intensity indicates the value of the ${\partial H_{3D-Har}^i}/{\partial {p_i}}$ result from corresponding $[p_i, L_{reg}, L_{dir}]$. Three axis represent the value of ${L_{reg}}$, ${p_i}$ and ${L_{dir}}$ respectively. Its counterpart ${\partial H_{3D}^i}/{\partial {p_i}}$ is shown in Fig.\ref{fig:fig3}(a) with same axis representation. 
Note: the gradient backpropagation from the classification part has no relationship with the regression part and direction estimation part which can be observed in Fig.\ref{fig:fig3}(a) (better view in its vertical view Fig.\ref{fig:fig3}(e)). In the case of 3D harmonic loss (better view in Fig.\ref{fig:fig3}(f)), the gradient from classification loss is suppressed by high regression loss (bad localization), leading to relatively low confidence, which brings mutual consistency between classification and localization. Additionally, the ${L_{dir}}$ gradually affect the gradient propagation, in order to achieve a globally unique optimization (${\partial H_{3D}^i}/{\partial {p_i}=0}$ only when ${p_i}=1$, ${L_{dir}}=0$ and ${L_{reg}}=0$).

\textbf{Proof-2: } 
The effectiveness analysis of 3D harmonic loss on regression part is derived as follows
\begin{equation}
    \begin{array}{l}\label{reg0}


\frac{{\partial L_{3D - Har}^i}}{{\partial \Delta {d_i}}} = {-e^{ - {L_{reg}}\left( {\Delta {d_i}} \right)}} {L_{cls}} \left( {\frac{{\partial {L_{reg}}\left( {\Delta {d_i}} \right)}}{{\Delta {d_i}}}} \right)\\
\,\,\,\,\,\,\,\,\,\,\,\,\,\,\,\,\,\,\,\,\,\, + \left( {1 + e^{-L_{cls}}} \right)\left( {\frac{{\partial {L_{reg}}\left( {\Delta {d_i}} \right)}}{{\partial \Delta {d_i}}}} \right)\\
\,\,\,\,\,\,\,\,\,\,\,\,\,\,\,\,\,\,\,\,\,\,\, + \frac{{{e^{ - {L_{reg}}\left( {\Delta {d_i}} \right)}} \left( {\frac{{\partial {L_{reg}}\left( {\Delta {d_i}} \right)}}{{\Delta {d_i}}}} \right)}}{2} \cdot {L_{dir}}

\end{array}
\end{equation}
The gradient back propagation from regression result is highly associated to $$\begin{array}{*{20}{c}}
{\Delta d_i}\\
 \uparrow \\
{regression}
\end{array},\begin{array}{*{20}{c}}
{L_{cls}}\\
 \uparrow \\
{classification}
\end{array},\begin{array}{*{20}{c}}
{{L_{dir}}}\\
 \uparrow \\
{direction\ estimation}
\end{array}$$

\textbf{Analysis-2: }
The Eq.\ref{reg0} outcomes are illustrated in Fig.\ref{fig:fig3}(d) using ten thousands data samples. The color intensity indicates the value of the ${\partial H_{3D-Har}^i}/{\partial {\Delta d_i}}$ result from corresponding $[\Delta d_i, L_{cls}, L_{dir}]$. Three-axis represent the value of ${\Delta d_i}$, ${L_{reg}}$, and ${L_{dir}}$ respectively. Its counterpart ${\partial H_{3D}^i}/{\partial {\Delta d_i}}$ is shown in Fig.\ref{fig:fig3}(c) with same axis representation. In previous 3D detection learning, the gradient backpropagation from the regression part is independent of the classification and direction estimation. Although our proposed method achieved the same regression result (same $\Delta d_i$), increasing classification loss will constantly restrain the gradient from maintaining the synchronous learning of classification and regression (better view in Fig.\ref{fig:fig3}(h)). The global unique optimization of ${\partial H_{3D}^i}/{\partial {\Delta d_i}=0}$ is achieved only when ${\Delta d_i}=0$, ${L_
cls}=0$ and ${L_{dir}}=0$).

\textbf{Proof-3: } 
The effectiveness analysis of 3D harmonic loss on the direction part is derived as follows. 
\begin{equation}
L_{dir}\left(p_i^{'} \right) =  \left( 1 - {p_i^{{'}^{gt}}} \right)\log \left( 1 - p_i^{'} \right) - p_i^{{'}^{gt}} log\left( p_i^{'} \right)
\end{equation}
Such that $$\frac{\partial {L_{dir}}\left( p_i^{'} \right)}{\partial p_i^{'}} = - \frac{p_i^{{'}^{gt}} - p_i^{'}}{\left( 1 - p_i^{'} \right)p_i^{'}} = \mu \left( i \right)$$ Based on the $p_i^{{'}^{gt}}$ status, the binary cross entropy loss is updated as follows
\begin{equation}
    \mu \left( i \right) = \left\{ \begin{array}{l}
 - \frac{1}{p_i^{'}},\,\,\,if\,\,\,p_i^{{'}^{gt}}= 1\\
 - \frac{1}{1 - p_i^{'}},\,\,\,if\,\,\,p_i^{{'}^{gt}} = 0
\end{array} \right.
\end{equation}
The gradient from the updated direction loss is as follows.
\begin{equation}
    \begin{array}{l}
\frac{\partial L_{3D-Har}^{i}}{\partial p_i^{'}} = \left( {1 - \frac{{{\beta _r} + {\beta _c}}}{2}} \right) \cdot \frac{\partial L_{3D-Har}^{i}}{\partial p_i^{'}}\\
\,\,\,\,\,\,\,\,\,\,\,\,\,\,\,\,\,\,\,\,\,\, = \left( {1 - \frac{{{\beta _r} + {\beta _c}}}{2}} \right) \cdot \left[ { - \frac{p_i^{{'}^{gt}}}{p_i^{'}} - \frac{{\left( {1 - p_i^{{'}^{gt}}} \right)}}{1 - p_i^{'}}} \right]
\end{array}
\end{equation}
The type of direction loss estimation is dependent on the $p_i^{{'}^{gt}}$ status, and it is derived as follows. 
\begin{equation}\label{dir0}
    \frac{\partial L_{3D - Har}^{i}}{\partial p_i^{'}} = \left\{ \begin{array}{l}
\left( {1 - \frac{{{{e^{ - {L_{cls}}} + {e^{ - {L_{reg}}}}}}}}{2}} \right)\left( { - \frac{1}{p_i^{'}}} \right),\,if\,p_i^{{'}^{gt}} = 1\\
\left( {1 - \frac{{{{e^{ - {L_{cls}}} + {e^{ - {L_{reg}}}}}}}}{2}} \right)\left( { - \frac{1}{1 - p_i^{'}}} \right),\,if\,p_i^{{'}^{gt}} = 0
\end{array} \right.
\end{equation}
The gradient backpropagation from the direction result is highly associated to $$\begin{array}{*{20}{c}}
{{L_{reg}}}\\
 \uparrow \\
{regression}
\end{array},\begin{array}{*{20}{c}}
{{L_{cls}}}\\
 \uparrow \\
{classification}
\end{array},\begin{array}{*{20}{c}}
p_i^{'}\\
 \uparrow \\
{direction\ estimation}
\end{array}$$

\textbf{Analysis-3: }
The Eq.\ref{dir0} outcomes are illustrated in Fig.\ref{fig:visdir} using ten thousands data samples. The color intensity indicates the value of the ${\partial H_{3D-Har}^{i}}/{\partial {p_i^{'}}}$ result from corresponding $[p_i^{'}, L_{cls}, L_{reg}]$. Three axis represent the value of ${p_i^{'}}$, ${L_{reg}}$, and ${L_{cls}}$, respectively. For instance, in Fig.\ref{fig:visdir}(a), given ${p_i^{{'}^{gt}}}=0$, the global unique optimization is ${\partial H_{3D}^{i}}/{\partial {p_i^{'}}=0}$ because when ${p_i^{'}}=0$, ${L_cls}=0$ and ${L_{reg}}=0$. Fig.\ref{fig:visdir}(b) shows the same paradigm of such intrinsic-consistency when ${p_i^{{'}^{gt}}}=1$.
}


	\begin{figure}[!htb]
		\centering
		\includegraphics[width=0.9\linewidth]{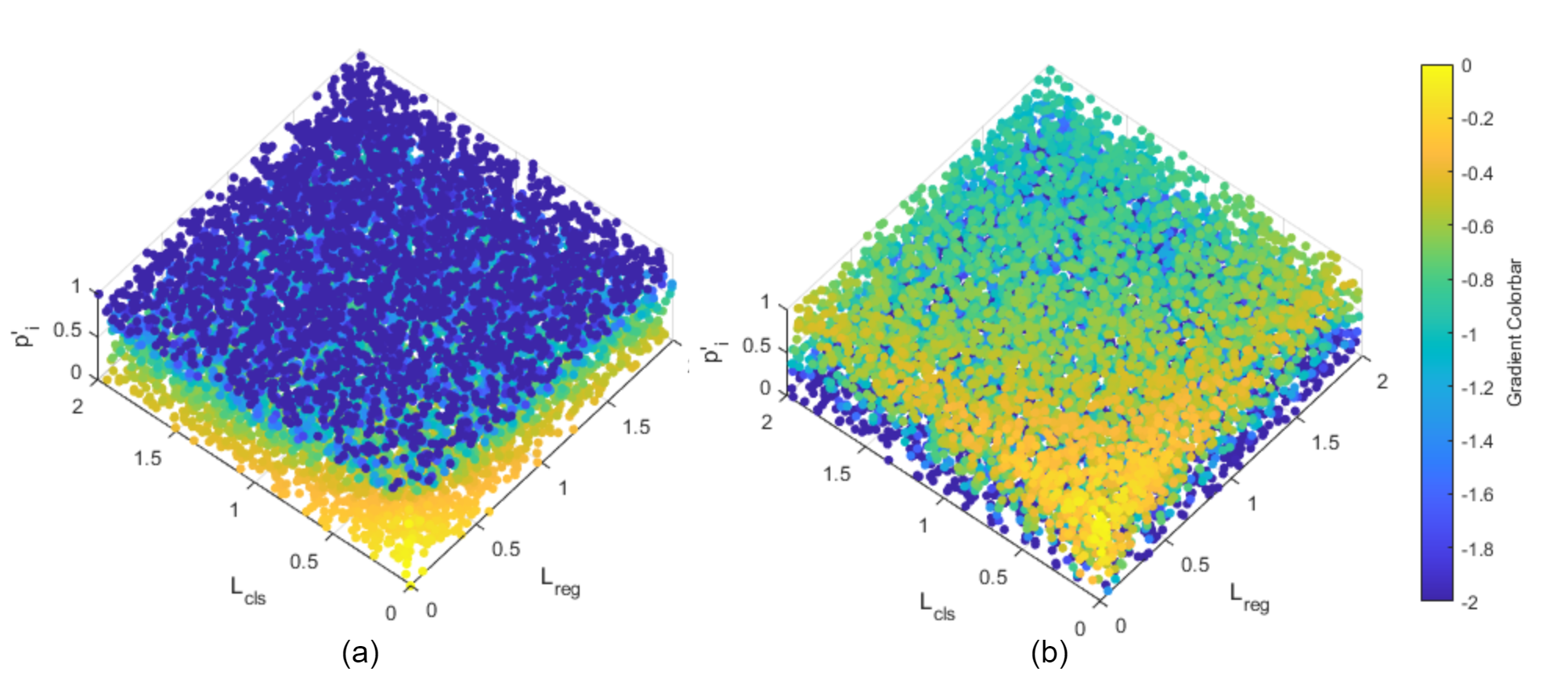}
		\caption{Visualization of the gradients from direction estimation part in our proposed 3D harmonic loss. (a) when $p_i^{{'}^{gt}}= 0$. (b) when $p_i^{{'}^{gt}}= 1$.}
		\label{fig:visdir}
	\end{figure}
\begin{table*}[!h]
\centering
\renewcommand{\arraystretch}{0.9}  
\setlength\tabcolsep{11pt}  
\centering
\resizebox{\textwidth}{!}{%
\begin{tabular}{|c|c|ccc|ccc|}
\hline
\multirow{2}{*}{\textbf{Method}} & \multirow{2}{*}{\textbf{Type / Modality}}  & \multicolumn{3}{c|}{\textbf{IoU threshold: 0.7}} & \multicolumn{3}{c|}{\textbf{IoU threshold: 0.5}} \\ \cline{3-8} 
&   & \multicolumn{1}{c|}{\textbf{Easy}} & \multicolumn{1}{c|}{\textbf{Mod.}} & \textbf{Hard} & \multicolumn{1}{c|}{\textbf{Easy}} & \multicolumn{1}{c|}{\textbf{Mod.}} & \textbf{Hard} \\ \hline \hline
 
 \textbf{PointPillar\cite{pp} ${\star}$} & one-stage / LiDAR    &  \multicolumn{1}{c|}{N/A} & \multicolumn{1}{c|}{87.70} & N/A & \multicolumn{1}{c|}{N/A} & \multicolumn{1}{c|}{N/A} & N/A \\ 
 \cdashline{1-8}[0.5pt/2pt]
\textbf{PointPillar\cite{pp} ${\dagger}$} & one-stage / LiDAR    &  \multicolumn{1}{c|}{92.09} & \multicolumn{1}{c|}{87.85} & 83.35 & \multicolumn{1}{c|}{95.72} & \multicolumn{1}{c|}{94.72} & 90.08  \\ 
\textbf{Harmonic PointPillar (Ours) ${\dagger}$} & one-stage / LiDAR   & \multicolumn{1}{c|}{94.07} & \multicolumn{1}{c|}{88.41} & 85.42 & \multicolumn{1}{c|}{95.98} & \multicolumn{1}{c|}{94.87} & 92.09  \\ 
\rowcolor{sblue}\textbf{${\Delta}$} & N/A    &\multicolumn{1}{c|}{\textbf{\textcolor{blue}{+1.98}}} & \multicolumn{1}{c|}{\textcolor{blue}{+0.56}} & \textbf{\textcolor{blue}{+2.07}} & \multicolumn{1}{c|}{\textcolor{blue}{+0.26}} & \multicolumn{1}{c|}{\textcolor{blue}{+0.15}} & \textbf{\textcolor{blue}{+2.01}}   \\ \hline \hline

\textbf{SECOND\cite{second} ${\star}$} & one-stage / LiDAR   & \multicolumn{1}{c|}{89.96} & \multicolumn{1}{c|}{87.07} &79.66  &  \multicolumn{1}{c|}{N/A} & \multicolumn{1}{c|}{N/A} &N/A   \\ 
\cdashline{1-8}[0.5pt/2pt]
\textbf{SECOND\cite{second} ${\dagger}$} & one-stage / LiDAR  & \multicolumn{1}{c|}{93.47} & \multicolumn{1}{c|}{88.96} & 86.23 & \multicolumn{1}{c|}{96.60} & \multicolumn{1}{c|}{95.27} & 92.60  \\ 
\textbf{Harmonic Second (Ours) ${\dagger}$} & one-stage / LiDAR   & \multicolumn{1}{c|}{95.41} & \multicolumn{1}{c|}{89.23} & 86.25 & \multicolumn{1}{c|}{98.96} & \multicolumn{1}{c|}{95.63} & 94.63  \\ 
\rowcolor{sblue}\textbf{${\Delta}$} & N/A   & \multicolumn{1}{c|}{\textbf{\textcolor{blue}{+1.94}}} & \multicolumn{1}{c|}{\textcolor{blue}{+0.27}} & \textcolor{blue}{+0.02} & \multicolumn{1}{c|}{\textbf{\textcolor{blue}{+2.36}}} & \multicolumn{1}{c|}{\textcolor{blue}{+0.36}} & \textbf{\textcolor{blue}{+2.03}}  \\ \hline \hline

\textbf{Point RCNN\cite{pointrcnn} ${\star}$} & two-stage / LiDAR  & \multicolumn{1}{c|}{N/A} & \multicolumn{1}{c|}{N/A} & N/A & \multicolumn{1}{c|}{N/A} & \multicolumn{1}{c|}{N/A} & N/A  \\ \cdashline{1-8}[0.5pt/2pt]
\textbf{Point RCNN\cite{pointrcnn} ${\dagger}$} & two-stage / LiDAR  & \multicolumn{1}{c|}{94.69} & \multicolumn{1}{c|}{88.53} & 88.14 & \multicolumn{1}{c|}{97.85} & \multicolumn{1}{c|}{94.31} & 94.15  \\ 
\textbf{Harmonic Point RCNN (Ours) ${\dagger}$} & two-stage / LiDAR  & \multicolumn{1}{c|}{94.97} & \multicolumn{1}{c|}{88.27} & 88.02 & \multicolumn{1}{c|}{98.45} & \multicolumn{1}{c|}{94.30} & 94.01  \\ 
\rowcolor{sblue}\textbf{$\Delta$} & N/A   & \multicolumn{1}{c|}{{\textcolor{blue}{+0.28}}} & \multicolumn{1}{c|}{{\textcolor{green}{-0.26}}} & {\textcolor{green}{-0.12}} & \multicolumn{1}{c|}{{\textcolor{blue}{+0.60}}} & \multicolumn{1}{c|}{{\textcolor{green}{-0.01}}} & {\textcolor{green}{-0.14}}  \\ \hline \hline

\textbf{Part-${A^2}$\cite{part2} ${\star}$} & two-stage / LiDAR   & \multicolumn{1}{c|}{90.42} & \multicolumn{1}{c|}{88.61} & 87.31 & \multicolumn{1}{c|}{N/A} & \multicolumn{1}{c|}{N/A} & N/A  \\ \cdashline{1-8}[0.5pt/2pt]
\textbf{Part-${A^2}$\cite{part2} ${\dagger}$} & two-stage / LiDAR  & \multicolumn{1}{c|}{92.78} & \multicolumn{1}{c|}{89.47} & 88.34 & \multicolumn{1}{c|}{96.95} & \multicolumn{1}{c|}{94.17} & 94.14 \\ 
\textbf{Harmonic Part-${A^2}$ (Ours) ${\dagger}$} & two-stage / LiDAR  & \multicolumn{1}{c|}{95.00} & \multicolumn{1}{c|}{90.14} & 88.38 & \multicolumn{1}{c|}{97.93} & \multicolumn{1}{c|}{95.41} & 94.02 \\ 
\rowcolor{sblue}\textbf{${\Delta}$} & N/A   & \multicolumn{1}{c|}{\textbf{\textcolor{blue}{+2.22}}} & \multicolumn{1}{c|}{{\textcolor{blue}{+0.67}}} & {\textcolor{blue}{+0.04}} & \multicolumn{1}{c|}{{\textbf{\textcolor{blue}{+0.98}}}} & \multicolumn{1}{c|}{\textbf{\textcolor{blue}{+1.24}}} & {\textcolor{green}{-0.12}}  \\ \hline 

\end{tabular}
}
\begin{tablenotes}
        \footnotesize
        \item ${\star}$: reported results in paper,  ${\dagger}$: our implementation on mmdetection3D\cite{mmdet3d}. N/A: not available or not applicable. ${\Delta}_{average}$ = +0.81. Emphases are highlighted in bold. 
      \end{tablenotes}
\caption{mAP evaluation of BEV object detection on car class of KITTI validation dataset }
\label{t1}
\end{table*}



\begin{table*}[!h]
\renewcommand{\arraystretch}{0.9}  
\setlength\tabcolsep{11pt}  
\centering
\resizebox{\textwidth}{!}{%
\begin{tabular}{|c|c|ccc|ccc|}
\hline
\multirow{2}{*}{\textbf{Method}} & \multirow{2}{*}{\textbf{Type / Modality}} & \multicolumn{3}{c|}{\textbf{IoU threshold: 0.7}} & \multicolumn{3}{c|}{\textbf{IoU threshold: 0.5}} \\ \cline{3-8} 
 &  & \multicolumn{1}{c|}{\textbf{Easy}} & \multicolumn{1}{c|}{\textbf{Mod.}} & \textbf{Hard} & \multicolumn{1}{c|}{\textbf{Easy}} & \multicolumn{1}{c|}{\textbf{Mod.}} & \textbf{Hard} \\ \hline \hline
 
 \textbf{PointPillar\cite{pp} $\star$} & one-stage / LiDAR  & \multicolumn{1}{c|}{N/A} & \multicolumn{1}{c|}{77.40} & N/A & \multicolumn{1}{c|}{N/A} & \multicolumn{1}{c|}{N/A} & N/A \\ \cdashline{1-8}[0.5pt/2pt]
\textbf{PointPillar\cite{pp} $\dagger$} & one-stage / LiDAR   & \multicolumn{1}{c|}{87.67} & \multicolumn{1}{c|}{76.44} & 73.27 & \multicolumn{1}{c|}{95.67} & \multicolumn{1}{c|}{94.41} & 89.88 \\ 
\textbf{Harmonic PointPillar (Ours) $\dagger$} & one-stage / LiDAR   & \multicolumn{1}{c|}{87.66} & \multicolumn{1}{c|}{77.76} & 73.44 & \multicolumn{1}{c|}{95.95} & \multicolumn{1}{c|}{94.72} & 90.12  \\
\rowcolor{sblue}\textbf{$\Delta$} & N/A  & \multicolumn{1}{c|}{ \textcolor{green}{-0.01}} & \multicolumn{1}{c|}{\textbf{\textcolor{blue}{+1.32}}} & \textcolor{blue}{+0.17} & \multicolumn{1}{c|}{\textcolor{blue}{+0.28}} & \multicolumn{1}{c|}{\textcolor{blue}{+0.31}} & \textcolor{blue}{+0.24}  \\ \hline \hline

\textbf{SECOND\cite{second} $\star$} & one-stage / LiDAR   & \multicolumn{1}{c|}{87.43} & \multicolumn{1}{c|}{76.48} &69.10
  &  \multicolumn{1}{c|}{N/A} & \multicolumn{1}{c|}{N/A} &N/A   \\ \cdashline{1-8}[0.5pt/2pt]
\textbf{SECOND\cite{second} $\dagger$} & one-stage / LiDAR   & \multicolumn{1}{c|}{89.58} & \multicolumn{1}{c|}{79.78} & 76.49 & \multicolumn{1}{c|}{96.56} & \multicolumn{1}{c|}{95.01} & 92.45  \\ 
\textbf{Harmonic Second (Ours) $\dagger$} & one-stage / LiDAR   & \multicolumn{1}{c|}{91.16} & \multicolumn{1}{c|}{79.68} & 76.06 & \multicolumn{1}{c|}{98.88} & \multicolumn{1}{c|}{95.36} & 92.56  \\
\rowcolor{sblue}\textbf{$\Delta$} & N/A   & \multicolumn{1}{c|}{\textbf{\textcolor{blue}{+1.58}}} & \multicolumn{1}{c|}{\textcolor{green}{-0.10}} & \textcolor{green}{-0.43} & \multicolumn{1}{c|}{\textbf{\textcolor{blue}{+2.32}}} & \multicolumn{1}{c|}{\textcolor{blue}{+0.35}} & \textcolor{blue}{+0.11}  \\ \hline \hline

\textbf{Point RCNN\cite{pointrcnn} ${\star}$} & two-stage / LiDAR  & \multicolumn{1}{c|}{88.88} & \multicolumn{1}{c|}{78.63} & 77.38 & \multicolumn{1}{c|}{N/A} & \multicolumn{1}{c|}{N/A} & N/A  \\ \cdashline{1-8}[0.5pt/2pt]
\textbf{Point RCNN\cite{pointrcnn} ${\dagger}$} & two-stage / LiDAR  & \multicolumn{1}{c|}{90.99} & \multicolumn{1}{c|}{80.20} & 77.93 & \multicolumn{1}{c|}{97.81} & \multicolumn{1}{c|}{94.20} & 93.83  \\ 
\textbf{Harmonic Point RCNN (Ours) ${\dagger}$} & two-stage / LiDAR  & \multicolumn{1}{c|}{91.77} & \multicolumn{1}{c|}{80.07} & 77.26 & \multicolumn{1}{c|}{98.41} & \multicolumn{1}{c|}{94.17} & 93.88  \\ 
\rowcolor{sblue}\textbf{$\Delta$} & N/A   & \multicolumn{1}{c|}{{\textcolor{blue}{+0.78}}} & \multicolumn{1}{c|}{{\textcolor{green}{-0.13}}} & {\textcolor{green}{-0.67}} & \multicolumn{1}{c|}{{\textcolor{blue}{+0.60}}} & \multicolumn{1}{c|}{{\textcolor{green}{-0.03}}} & {\textcolor{blue}{+0.05}}  \\ \hline \hline

\textbf{Part-$A^2$\cite{part2} $\star$} & two-stage / LiDAR   & \multicolumn{1}{c|}{ 89.47} & \multicolumn{1}{c|}{79.47} & 78.54 & \multicolumn{1}{c|}{N/A} & \multicolumn{1}{c|}{N/A} & N/A  \\ \cdashline{1-8}[0.5pt/2pt]
\textbf{Part-$A^2$\cite{part2} $\dagger$} & two-stage / LiDAR   & \multicolumn{1}{c|}{91.81} & \multicolumn{1}{c|}{82.35} & 80.16 & \multicolumn{1}{c|}{96.91} & \multicolumn{1}{c|}{94.09} & 93.95  \\ 
\textbf{Harmonic Part-$A^2$ (Ours) $\dagger$} & two-stage / LiDAR   & \multicolumn{1}{c|}{91.92} & \multicolumn{1}{c|}{82.43} & 80.07 & \multicolumn{1}{c|}{97.92} & \multicolumn{1}{c|}{94.01} & 93.85  \\ 

\rowcolor{sblue}\textbf{$\Delta$} & N/A & \multicolumn{1}{c|}{{\textcolor{blue}{+0.11}}} & \multicolumn{1}{c|}{{\textcolor{blue}{+0.08}}} & {\textcolor{green}{-0.09}} & \multicolumn{1}{c|}{\textbf{\textcolor{blue}{+1.01}}} & \multicolumn{1}{c|}{{\textcolor{green}{-0.08}}} & {\textcolor{green}{-0.10}}  \\ \hline 
\end{tabular}}

\begin{tablenotes}
        \footnotesize
        \item $\star$: reported results in paper,  $\dagger$: our implementation on mmdetection3D\cite{mmdet3d}. N/A: not available or not applicable. ${\Delta}_{average}$ = +0.32. Emphases are highlighted in bold.
      \end{tablenotes}

\caption{mAP evaluation of 3D object detection on car class of KITTI validation dataset }

\label{t2}
\end{table*}

\begin{table*}[!h]
\renewcommand{\arraystretch}{0.8}  
\setlength\tabcolsep{11pt}  
\centering
\resizebox{0.9\textwidth}{!}{%
\begin{tabular}{|c|c|c|ccc|}
\hline
\multirow{2}{*}{\textbf{Method}} & \multirow{2}{*}{\textbf{Source}} & \multirow{2}{*}{\textbf{Type / Modality}}& \multicolumn{3}{c|}{\textbf{Car (IoU threshold:0.7)}} \\ \cline{4-6} 
 &  &  &   \multicolumn{1}{c|}{\textbf{Easy}} & \multicolumn{1}{c|}{\textbf{Mod.}} & \textbf{Hard}  \\ \hline \hline

  \textbf{F-PointNet\cite{fpointnet} } & CVPR 2018 & fusion / LiDAR+camera   & \multicolumn{1}{c|}{91.17} & \multicolumn{1}{c|}{\textcolor{syellow}{84.67}} & \textcolor{syellow}{74.77} \\
  \textbf{PointPainting\cite{ppint} } & CVPR 2020 & fusion / LiDAR+camera   & \multicolumn{1}{c|}{92.45} & \multicolumn{1}{c|}{88.11} & 83.36 \\
  \textbf{CLOCs\cite{clocs} } & IROS 2020 & fusion / LiDAR+camera   & \multicolumn{1}{c|}{91.16} & \multicolumn{1}{c|}{88.23} & 82.63  \\
   \textbf{StructuralIF\cite{IF} } & CVIU 2021 & fusion / LiDAR+camera   & \multicolumn{1}{c|}{91.78} & \multicolumn{1}{c|}{88.38} & 85.67 
 \\ 
 \hline
 
 \textbf{Point-RCNN\cite{pointrcnn} } & CVPR 2019 & two-stage / LiDAR   & \multicolumn{1}{c|}{92.13} & \multicolumn{1}{c|}{87.39} & 82.72 \\
\textbf{PI-RCNN\cite{pircnn} } & AAAI 2020 & two-stage / LiDAR   & \multicolumn{1}{c|}{91.44} & \multicolumn{1}{c|}{\textcolor{syellow}{85.81}} & \textcolor{syellow}{81.00} \\
\textbf{Part-$A^2$\cite{part2} } & TPAMI 2021 & two-stage / LiDAR   & \multicolumn{1}{c|}{91.70} & \multicolumn{1}{c|}{87.79} & 84.61 \\
\textbf{Voxel-RCNN\cite{voxelrcnn} } & AAAI 2021 & two-stage / LiDAR   & \multicolumn{1}{c|}{94.85} & \multicolumn{1}{c|}{88.83} & 86.13 \\
\textbf{EQ-PVRCNN\cite{eqpvrcnn} } & CVPR 2022 & two-stage / LiDAR   & \multicolumn{1}{c|}{94.55} & \multicolumn{1}{c|}{89.09} & 86.42 \\
 \hline

\textbf{3DSSD\cite{3dssd}} & CVPR 2020 & one-stage / LiDAR   & \multicolumn{1}{c|}{92.66} & \multicolumn{1}{c|}{89.02} & 85.86 \\

\textbf{Point-GNN\cite{pointgnn}} & CVPR 2020 & one-stage / LiDAR   & \multicolumn{1}{c|}{93.11} & \multicolumn{1}{c|}{89.17} & 83.90\\

\textbf{TANet\cite{tanet} } & AAAI 2020 & one-stage / LiDAR    & \multicolumn{1}{c|}{91.58} & \multicolumn{1}{c|}{\textcolor{syellow}{86.54}} & \textcolor{syellow}{81.19} \\

\textbf{VoxSet\cite{voxel-set} } & CVPR 2022 & one-stage / LiDAR   & \multicolumn{1}{c|}{92.70} & \multicolumn{1}{c|}{89.07} & 86.29 \\

\cdashline{1-6}[0.5pt/2pt]
\textbf{PointPillar\cite{pp} } & CVPR 2019 & one-stage / LiDAR   & \multicolumn{1}{c|}{\textcolor{syellow}{90.07}} & \multicolumn{1}{c|}{\textcolor{syellow}{86.56}} & 82.81
\\ 
\textbf{Harmonic PointPillar} & Ours  & one-stage / LiDAR   & \multicolumn{1}{c|}{90.89} & \multicolumn{1}{c|}{87.28} & 82.54
\\
\rowcolor{sblue} \textbf{${\Delta}$} & N/A  & N/A    & \multicolumn{1}{c|}{{\textbf{\textcolor{blue}{+0.82}}}} & \multicolumn{1}{c|}{\textbf{\textcolor{blue}{+0.72}}} & {\textcolor{green}{-0.27}}  \\ \hline

\end{tabular}}

\begin{tablenotes}
        \footnotesize
        \item Results of listed works were extracted from KITTI BEV test benchmark\cite{kt} (Date: 14 August 2022). N/A: not applicable. ${\Delta}_{average}$ = +0.42. Results worse than Harmonic PointPillar (Ours) are colored in \textcolor{syellow}{orange}. Check our submitted result at \url{https://www.cvlibs.net/datasets/kitti/eval_object_detail.php?&result=cf021462bb1955480c0c5ebe6c1756545bf98566}.
      \end{tablenotes}

\caption{mAP evaluation of BEV object detection on car class of KITTI test benchmark}

\label{ttest}
\end{table*}

\begin{table}[!h]
\renewcommand{\arraystretch}{0.9}  
\setlength\tabcolsep{2.2pt}  
\centering
\resizebox{0.9\textwidth}{!}{%
\begin{tabular}{|c|ccc|ccc|}
\hline
\multirow{2}{*}{\textbf{Method}}  & \multicolumn{3}{c|}{\textbf{BEV Car}} & \multicolumn{3}{c|}{\textbf{3D Car}} \\ \cline{2-7} 
  & \multicolumn{1}{c|}{\textbf{Easy}} & \multicolumn{1}{c|}{\textbf{Mod.}} & \textbf{Hard} & \multicolumn{1}{c|}{\textbf{Easy}} & \multicolumn{1}{c|}{\textbf{Mod.}} & \textbf{Hard} \\ \hline \hline
 
\textbf{PointPillar\cite{pp} }  & \multicolumn{1}{c|}{64.77} & \multicolumn{1}{c|}{54.76} & 54.77 & \multicolumn{1}{c|}{64.40} & \multicolumn{1}{c|}{54.39} & 54.43 \\ 
\textbf{Harmonic PointPillar (Ours) }    & \multicolumn{1}{c|}{66.89} & \multicolumn{1}{c|}{54.46} & 54.48 & \multicolumn{1}{c|}{64.10} & \multicolumn{1}{c|}{54.10} & 54.13  \\
\rowcolor{sblue}\textbf{$\Delta$}  & \multicolumn{1}{c|}{ \textbf{\textcolor{blue}{+2.12}}} & \multicolumn{1}{c|}{\textcolor{green}{-0.30}} & \textcolor{green}{-0.29} & \multicolumn{1}{c|}{\textcolor{green}{-0.30}} & \multicolumn{1}{c|}{\textcolor{green}{-0.29}} & \textcolor{green}{-0.30}  \\ \hline \hline

\textbf{SECOND\cite{second} }   & \multicolumn{1}{c|}{67.10} & \multicolumn{1}{c|}{54.65} & 57.04 & \multicolumn{1}{c|}{66.36} & \multicolumn{1}{c|}{54.04} & 54.05  \\ 
\textbf{Harmonic Second (Ours) }    & \multicolumn{1}{c|}{67.12} & \multicolumn{1}{c|}{54.65} & 57.06 & \multicolumn{1}{c|}{66.42} & \multicolumn{1}{c|}{54.11} & 54.13  \\
\rowcolor{sblue}\textbf{$\Delta$}   & \multicolumn{1}{c|}{{\textcolor{blue}{+0.02}}} & \multicolumn{1}{c|}{N/A} & \textcolor{blue}{+0.06} & \multicolumn{1}{c|}{{\textcolor{blue}{+0.07}}} & \multicolumn{1}{c|}{N/A} & \textcolor{blue}{+0.08}  \\ \hline
\end{tabular}}

\begin{tablenotes}
        \footnotesize
        \item Results are from our implementation on mmdetection3D\cite{mmdet3d} following DAIR-V2X toolkit\cite{dair}. N/A: not available or not applicable. IoU threshold: 0.7. ${\Delta}_{average}$ = +0.07. Emphases are highlighted in bold.
      \end{tablenotes}

\caption{mAP evaluation of 3D/BEV object detection on car class of DAIR-V2X-I dataset}

\label{tdair}
\end{table}

\section{Experiments and Analysis}\label{sec4}

	\begin{figure*}[!htb]
		\centering
		\includegraphics[width=\linewidth]{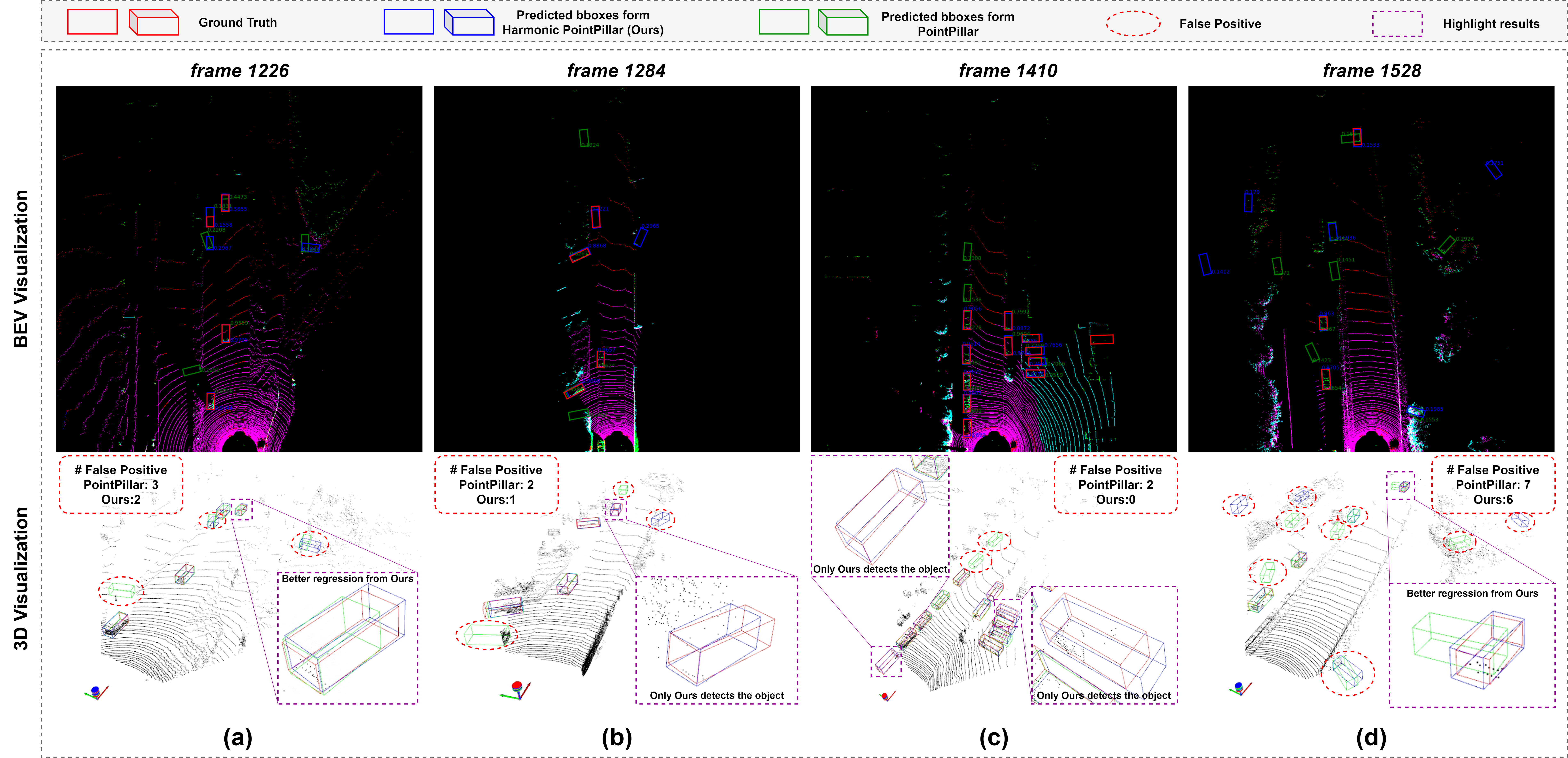}
		\caption{Qualitative analysis of overall 3D detection performance. Predicted bboxes from  {\textcolor{green}{Pointpillar (baseline) (green bboxes)}} {\cite{pp}} and predicted bboxes from {\textcolor{blue}{Harmonic Pointpillar (Ours) (blue bboxes)}} are visualized in same frames. {\textcolor{red}{Ground Truths (red bboxes)}} are also drawn for qualitative check. Harmonic PointPillar (Ours) shows better recall rate and localization accuracy with less false positive than PointPillar (baseline). }
		\label{fig:vis1}
	\end{figure*}

	\begin{figure*}[!htb]
		\centering
		\includegraphics[width=\linewidth]{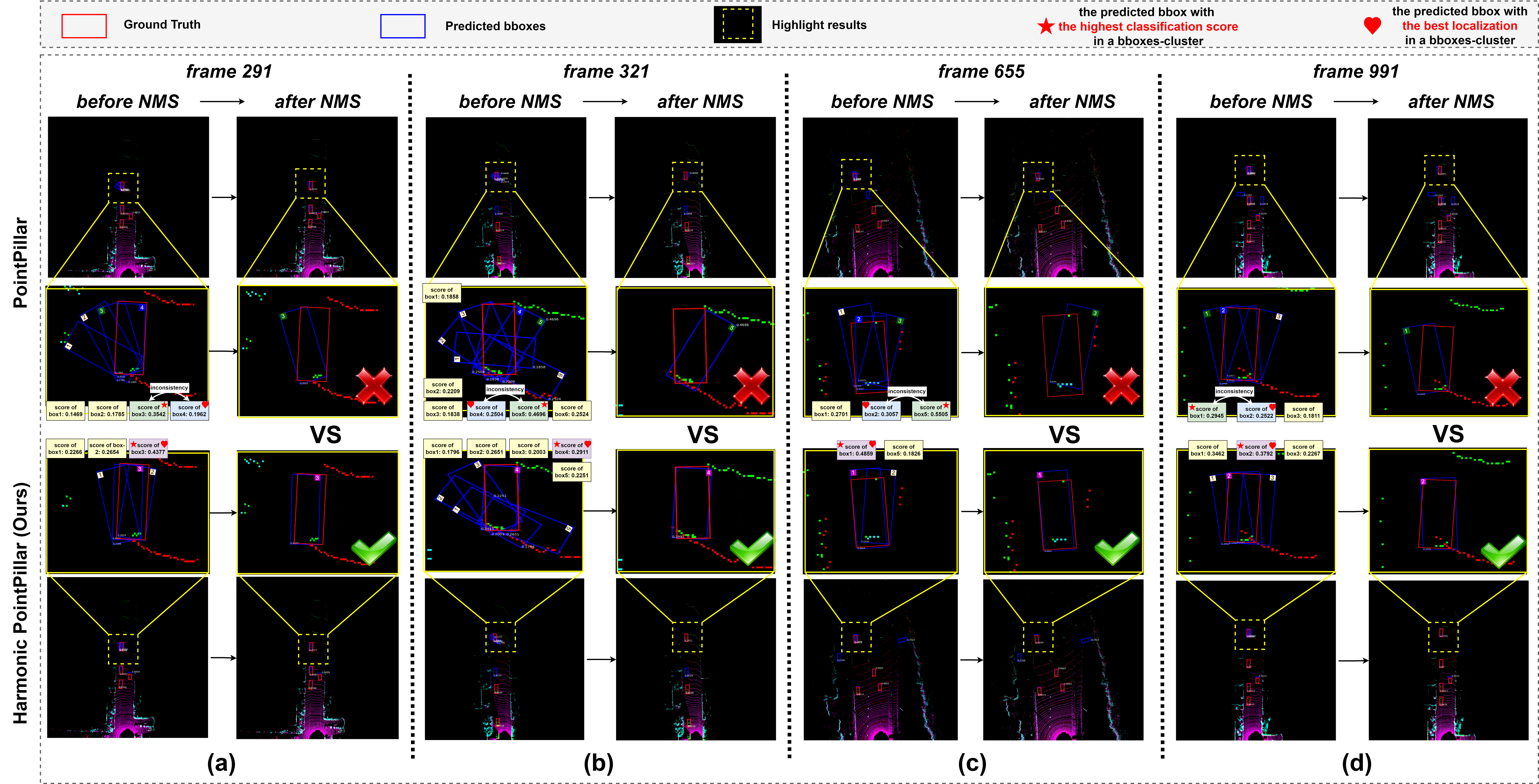}
		\caption{Qualitative analysis of inconsistent/consistent 3D detection. For better view, the results are in BEV visualization (zoom in for detail check). PointPillar (baseline) \cite{pp} suffers from inconsistency problem in 3D detection, while Harmonic PointPillar (Ours) shows a great robustness on keeping predictions consistent.}
		\label{fig:vis2}
	\end{figure*}



\subsection{Dataset and evaluation metrics}
\mm{The proposed 3D harmonic loss method performance is evaluated based on KITTI \cite{kt} dataset and DAIR-V2X-I dataset \cite{dair}. Both datasets offer LiDAR pointcloud data and 3D object annotations. 

KITTI dataset has 7481 training frames and 7518 test frames and following previous work \cite{pp,second,pointrcnn,part2} separated the 7481 training frames into the training dataset (3712 frames) and validation dataset (3769 frames). Accordingly, mean average precision (mAP) with 40 recall positions and Average Orientation Similarity (AOS) are adopted as evaluation metrics for detection accuracy analysis. 

DAIR-V2X \cite{dair} dataset enables DAIR-V2X-I sub dataset for infrastructure-based 3D object detection experiments. It contains 10k lidar pointcloud frames scanned from LiDAR installed on the infrastructure side. About 493k 3D objects are annotated in these frames. Like the KITTI dataset, DAIR-V2X-I has three classes such as car, pedestrian and cyclist. Following the DAIR official toolkit and their experiments \cite{dair}, we transfer the DAIR-V2X-I dataset into the KITTI data format and adopt the same evaluation metrics as the KITTI dataset.


\subsection{Implementation}
In this work, we conducted experiments on a server with a single NVIDIA GeForce RTX 2080Ti GPU. On KITTI dataset, four widely used models (one-stage detectors: PointPillar\cite{pp} and SECOND\cite{second}, two-stage detectors: PointRCNN\cite{pointrcnn} and Part-${A^{2}}$\cite{part2}) are adopted as baselines for estimating proposed model effectiveness. We re-implemented and trained these models using mmdetection3D \cite{mmdet3d} platform. Besides applying our proposed 3D harmonic loss, these models were trained with their original training settings and parameters. Our models are named Harmonic PointPillar, Harmonic SECOND, Harmonic PointRCNN, and Harmonic Part-${A^{2}}$. The post-processing is kept the same as the baselines in the evaluation stage. We also submitted the results of Harmonic PointPillar on the KITTI test dataset to the KITTI official benchmark. We compared the performance of our model with PointPillar (baseline), and other models \cite{fpointnet,ppint,clocs,IF,pircnn,voxelrcnn,eqpvrcnn,3dssd,pointgnn,tanet,voxel-set}, including fusion-based, two-stage lidar-based and one-stage lidar-based methods. PointPillar \cite{pp} and SECOND \cite{second}, two widely used one-stage detectors, are taken as baselines to assess the proposed model performance using the DAIR-V2X-I dataset. We implemented them on mmdetection3D \cite{mmdet3d} following the DAIR-V2X toolkit with original training parameters and evaluation settings. For a fair comparison, the only difference between our models and baselines is that our models adopt the proposed 3D harmonic loss.

\subsection{Quantitative analysis}
Experimental results with thorough quantitative analysis are reported below.


\textbf{Car detection} is an essential element for ITS applications (e.g. V2V and V2X). We test our method's performance from on-vehicle settings and roadside deployment. Tab.\ref{t1} and Tab.\ref{t2} show the mAP of car detection with IoU=0.7 and 0.5, and our proposed method achieves better average mAP values than baselines. \msm{Especially on BEV detection, our models achieved notables mAP rate (at least 0.02\% and at most 2.36\% beyond SECOND, at least 0.15\% and at most 2.07\% beyond PointPillar). }
We also submitted our model Harmonic PointPillar to the KITTI official test benchmark (see Tab.\ref{ttest}) and our method's time efficiency (ref to Tab.\ref{time}) makes it popular for industrial applications. Such as, our method optimizes the baseline PointPillar by 0.82\% improvement on Easy samples and 0.72\% improvement on Moderate samples, with only a 0.27\% decrease on Hard samples. Since our method focuses on balancing and harmonizing the gradient from different parts, for hard samples, usually with very sparse points scanned from objects, the classification confidence is low, suppressing the regression part and causing the mAP drop. On the other side, extremely hard samples (outliers) may affect the stability of the model because they always bring large gradients variant. Tab.\ref{tdair} shows evaluation results on the DAIR-V2X-I dataset. Our model achieves a minor average improvement (0.07\%). Since most current LiDAR-based 3D car detection models were designed and experimented with from an on-vehicle LiDAR view. Our future work will be to design better 3D detection methods tailored for on-infrastructure LiDAR.

\textbf{Direction estimation:} We use the average orientation similarity (AOS) index to evaluate the performance of 3D direction. AOS evaluation under different IoU threshold (0.7 and 0.5) are shown in Tab. \ref{t3aos}. A higher AOS value represents better direction estimation for 3D objects. Our proposed model achieved an average improvement of 0.26\% from PointPillar to Harmonic PointPillar and 0.76\% from SECOND to Harmonic SECOND. Specifically, the proposed strategy helps vanilla models to gain a large improvement on easy-level objects (at least 0.35\% and at most 1.99\% improvement under 0.7 IoU threshold; at least 0.22\% and at most 1.76\% improvement under 0.5 IoU threshold) and hard-level objects (at least 1.50\% and at most 1.52\% improvement under 0.5 IoU threshold). The results prove that our proposed method can better estimate object direction.}
\begin{table}[!h]
\renewcommand{\arraystretch}{0.9}  
\setlength\tabcolsep{2.2pt}  
\centering

\caption{Average Orientation Similarity (AOS) evaluation of one-stage 3D/BEV object detection on car class of KITTI validation dataset and KITTI test benchmark}

\resizebox{0.9\textwidth}{!}{%
\begin{tabular}{|c|ccc|ccc|}
\hline
\multirow{2}{*}{\textbf{Method}} &
\multicolumn{3}{c|}{\textbf{IoU threshold: 0.7}} &
\multicolumn{3}{c|}{\textbf{IoU threshold: 0.5}}  \\ \cline{2-7} 
 & \multicolumn{1}{c|}{\textbf{Easy}} & \multicolumn{1}{c|}{\textbf{Mod.}} & \textbf{Hard} & \multicolumn{1}{c|}{\textbf{Easy}} & \multicolumn{1}{c|}{\textbf{Mod.}} & \textbf{Hard}  \\ \hline \hline
 
 \textbf{SECOND\cite{second}$\diamond$}  & \multicolumn{1}{c|}{96.27} & \multicolumn{1}{c|}{92.13} & 88.94 & \multicolumn{1}{c|}{96.50} & \multicolumn{1}{c|}{94.95} & 91.84 \\ 
\textbf{Harmonic SECOND (Ours)$\diamond$}   & \multicolumn{1}{c|}{98.26} & \multicolumn{1}{c|}{92.00} & 88.56 & \multicolumn{1}{c|}{98.69} & \multicolumn{1}{c|}{94.77} & 93.36 \\ 
\rowcolor{sblue}\textbf{${\Delta}$}    & \multicolumn{1}{c|}{ \textbf{\textcolor{blue}{+1.99}}} & \multicolumn{1}{c|}{\textcolor{green}{-0.13}} & \textcolor{green}{-0.38} & \multicolumn{1}{c|}{\textbf{\textcolor{blue}{+1.76}}} & \multicolumn{1}{c|}{\textcolor{green}{-0.18}} & \textbf{\textcolor{blue}{+1.52}} \\ \hline \hline

\textbf{PointPillar\cite{pp}$\diamond$}  & \multicolumn{1}{c|}{95.31} & \multicolumn{1}{c|}{91.42} & 86.51 & \multicolumn{1}{c|}{95.71} & \multicolumn{1}{c|}{94.38} & 89.33 \\ 
\textbf{Harmonic PointPillar (Ours)$\diamond$}   & \multicolumn{1}{c|}{95.66} & \multicolumn{1}{c|}{91.33} & 86.21 & \multicolumn{1}{c|}{95.93} & \multicolumn{1}{c|}{94.28} & 90.83 \\
\rowcolor{sblue}\textbf{${\Delta}$}    & \multicolumn{1}{c|}{ \textcolor{blue}{+0.35}} & \multicolumn{1}{c|}{\textcolor{green}{-0.09}} & \textcolor{green}{-0.30} & \multicolumn{1}{c|}{\textcolor{blue}{+0.22}} & \multicolumn{1}{c|}{\textcolor{green}{-0.10}} & \textbf{\textcolor{blue}{+1.50}} \\ \hline \hline

\textbf{PointPillar\cite{pp}{$\ast$}}  & \multicolumn{1}{c|}{93.84} & \multicolumn{1}{c|}{90.70} & 87.47 & \multicolumn{1}{c|}{N/A} & \multicolumn{1}{c|}{N/A} & N/A \\ 
\textbf{Harmonic PointPillar (Ours){$\ast$}}   & \multicolumn{1}{c|}{94.23} & \multicolumn{1}{c|}{90.78} & 87.42 & \multicolumn{1}{c|}{N/A} & \multicolumn{1}{c|}{N/A} & N/A \\
\rowcolor{sblue}\textbf{${\Delta}$}    & \multicolumn{1}{c|}{ \textcolor{blue}{+0.39}} & \multicolumn{1}{c|}{\textcolor{blue}{+0.08}} & \textcolor{green}{-0.05} & \multicolumn{1}{c|}{N/A} & \multicolumn{1}{c|}{N/A} & N/A \\ \hline

\end{tabular}
}

\begin{tablenotes}
    \footnotesize
    \item {$\diamond$}: results on KITTI validation dataset. {$\ast$}: results on KITTI test benchmark. N/A: not applicable. Emphases are highlighted in bold.
  \end{tablenotes}
	
\label{t3aos}
\end{table}

\mm{\textbf{Vulnerable road users detection:} Besides car detection, detecting vulnerable road users (pedestrians and cyclists) provides additional security monitoring value to the V2X applications. From our observation and analysis, unlike vehicles, the scanned pointcloud shapes of pedestrians and cyclists are more irregular with different postures, which creates severe difficulty in model optimization. Tab.\ref{t4pedcyc} lists the mAP comparison of detecting pedestrians and cyclists under the official 0.5 IoU threshold. The proposed method achieved a better synchronous learning rate for classification, localization and direction estimation (related to the object shape). Compared to the base models, our model markedly boosts the accuracy of pedestrian and cyclist detection at most mAP improvement of 1.39\%, and 0.49\%, respectively. It shows that our proposed method is highly reliable in promoting the 3D detection of vulnerable traffic objects.}
\begin{table}[!h]
\renewcommand{\arraystretch}{0.9}  
\setlength\tabcolsep{2.5pt}  
\centering
\resizebox{0.9\textwidth}{!}{%
\begin{tabular}{|c|ccc|ccc|}
\hline
\multirow{2}{*}{\textbf{Method}} &
\multicolumn{3}{c|}{\textbf{3D Pedestrian}}  & \multicolumn{3}{c|}{\textbf{3D Cyclist}}  \\ \cline{2-7} 
 & \multicolumn{1}{c|}{\textbf{Easy}} & \multicolumn{1}{c|}{\textbf{Mod.}} & \textbf{Hard} & \multicolumn{1}{c|}{\textbf{Easy}} & \multicolumn{1}{c|}{\textbf{Mod.}} & \textbf{Hard} 
 \\ \hline \hline

 \textbf{SECOND\cite{second}}  & \multicolumn{1}{c|}{61.59} & \multicolumn{1}{c|}{54.27} & 48.03 & \multicolumn{1}{c|}{83.52} & \multicolumn{1}{c|}{65.04} & 60.98 \\ 
\textbf{Harmonic SECOND (Ours)}   & \multicolumn{1}{c|}{62.41} & \multicolumn{1}{c|}{55.46} & 48.93  & \multicolumn{1}{c|}{83.78} & \multicolumn{1}{c|}{64.90} &61.07 \\ 
\rowcolor{sblue}\textbf{${\Delta}$}    & \multicolumn{1}{c|}{ \textbf{\textcolor{blue}{+0.82}}} & \multicolumn{1}{c|}{\textbf{\textcolor{blue}{+1.19}}} & \textbf{\textcolor{blue}{+0.90}}  & \multicolumn{1}{c|}{\textcolor{blue}{+0.26}} & \multicolumn{1}{c|}{\textcolor{green}{-0.14}} & \textcolor{blue}{+0.09}  \\ \hline \hline
 
\textbf{PointPillar\cite{pp}}  & \multicolumn{1}{c|}{52.49} & \multicolumn{1}{c|}{46.47} & 41.52 & \multicolumn{1}{c|}{75.94} & \multicolumn{1}{c|}{59.19} & 55.89 \\ 
\textbf{Harmonic PointPillar (Ours)}   & \multicolumn{1}{c|}{52.31} & \multicolumn{1}{c|}{46.31} & 41.33  & \multicolumn{1}{c|}{77.33} & \multicolumn{1}{c|}{60.49} &56.48 \\ 
\rowcolor{sblue}\textbf{${\Delta}$}    & \multicolumn{1}{c|}{ \textcolor{green}{-0.18}} & \multicolumn{1}{c|}{\textcolor{green}{-0.16}} & \textcolor{green}{-0.21}  & \multicolumn{1}{c|}{\textbf{\textcolor{blue}{+1.39}}} & \multicolumn{1}{c|}{\textbf{\textcolor{blue}{+1.30}}} & \textcolor{blue}{+0.59}  \\ \hline 

\end{tabular}

}

\begin{tablenotes}
    \footnotesize
    \item Results are from our implementation on mmdetection3D \cite{mmdet3d}. IoU threshold: 0.5. ${\Delta}_{average}$ = +0.49. Emphases are highlighted in bold.
  \end{tablenotes}

\caption{mAP evaluation of 3D object detection on pedestrian/cyclist class of KITTI validation dataset }
\label{t4pedcyc}
\end{table}

\mm{\textbf{Time efficiency} of 3D detection matters in V2X applications. Roadside units (RSU) enable fast 3D detection models to catch vehicle signals in real-time. Tab.\ref{time} shows the runtime comparison of different state-of-the-art 3D detectors. On average, our method, same as PointPillar \cite{pp}, is at least 1.5X times fast in speed than other methods since it works as a training optimizer, and does not cause a delay in detection inference. This measurement certifies that our method is time-friendly.}

\begin{table}[h!]
\renewcommand{\arraystretch}{1}  
\setlength\tabcolsep{2.2pt}  
\centering
\resizebox{0.9\textwidth}{!}{%
\begin{tabular}{|c|c|c|c|}
\hline
\multirow{1}{*}{\textbf{Method}} & \multirow{1}{*}{\textbf{Source}}  & \multirow{1}{*}{\textbf{Type / Modality}} &  \multirow{1}{*}{\textbf{Speed (Hz)}}  \\ \hline \hline 

\textbf{Point-RCNN\cite{pointrcnn}} & CVPR 2019 & two-stage / LiDAR & 2.7   \\

\textbf{Part-$A^2$\cite{part2}} & TPAMI 2021 & two-stage / LiDAR & 9.5   \\

\textbf{CenterPoint\cite{centerpoint}} & CVPR 2021 &two-stage / LiDAR & 39.2\\

\hline

\textbf{SECOND\cite{second}} & Sensors 2018 & one-stage / LiDAR & 18.0   \\

\textbf{3DSSD\cite{3dssd}} & CVPR 2020 & one-stage / LiDAR & 10.9    \\

\textbf{Point-GNN\cite{pointgnn}} & CVPR 2020 & one-stage / LiDAR & 3.3   \\

\textbf{TANet\cite{tanet} } & AAAI 2020 & one-stage / LiDAR & 29.4    \\

\textbf{VoxSet\cite{voxel-set} } & CVPR 2022 & one-stage / LiDAR & 24.2    \\

\cdashline{1-4}[0.5pt/2pt]
\textbf{PointPillar\cite{pp} } & CVPR 2019 & one-stage / LiDAR & \textbf{43.1}   
\\ 
\textbf{Harmonic PointPillar} & Ours & one-stage / LiDAR & \textbf{43.1}  
 \\ \hline

\end{tabular}}

\begin{tablenotes}
\centering
        \footnotesize
        \item Inference speed was tested on Pytorch with single GPU 2080Ti.
      \end{tablenotes}

\caption{Average runtime comparison of 3D/BEV object detection}

\label{time}
\end{table}


\subsection{Qualitative analysis}
\begin{figure}[!htb]
		\centering
		\includegraphics[width=1.0\linewidth]{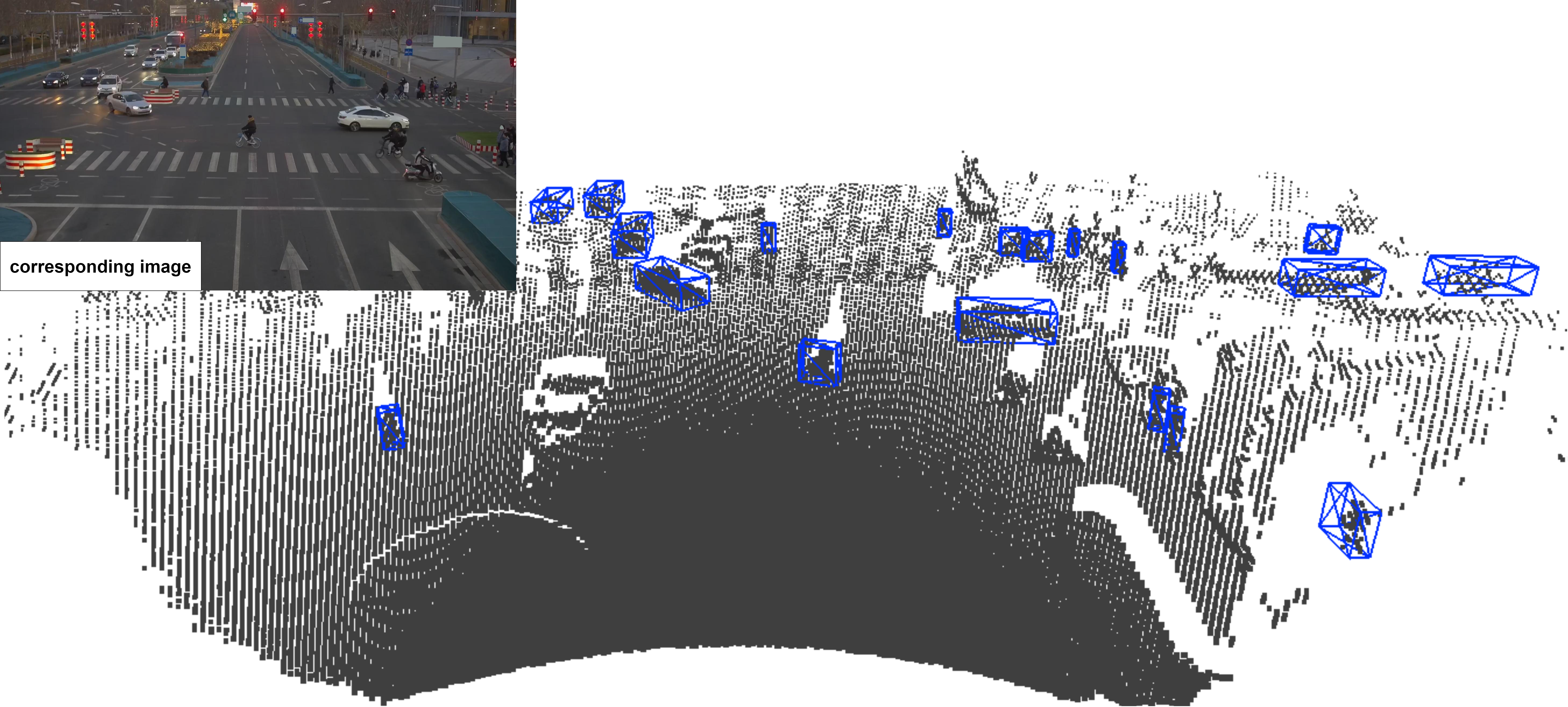}
		\caption{Qualitative example of on-infrastructure LiDAR-based 3D object detection (detection results ({\textcolor{blue}{blue bboxes}}) are made by our proposed method: Harmonic PointPillar).}
		\label{fig:deploy}
	\end{figure}

 \mm{Visualized results with detailed qualitative analysis are reported below.

\textbf{Overall performance:} Fig.\ref{fig:vis1} shows comparison of overall performance of 3D detection. In the examples of Fig.\ref{fig:vis1}(a) and Fig.\ref{fig:vis1}(d), our model (Harmonic PointPillar) achieves better localization accuracy than baseline model (PointPillar). In Fig.\ref{fig:vis1}(b) and Fig.\ref{fig:vis1}(c), our model detected more valid objects which are missed by the baseline models. Furthermore, our model's false positive (FP) ratio is less than the baseline model in all example frames. 

\textbf{Dealing with inconsistency problem:} In order to further verify the effectiveness of the proposed method in solving inconsistency problems in 3D detection, we show a more detailed qualitative visualization in Fig.\ref{fig:vis2} (zoom in for detail check). The baseline model (PointPillar) has not succeeded in predicting the targets in all example frames because of the inconsistency between classification and localization. Consequently, our model achieved outstanding robustness in preserving consistent 3D detection, resulting in the best predictions in all example frames. It confirms that our method can be used to build a task-consistent 3D detector.

\subsection{Simulations on Realistic Deployment}
Based on the experience from previous work \cite{ppdeploy}, we converted PyTorch-style Harmonic PointPillar into TensorRT-format. The tensorRT-format model is deployed on Jetson Xavier TX with float16 quantization techniques, and the same experiments as on PC were repeated on Jetson Xavier TX. A notable 2x-speed (75.4Hz on Jetson Xavier TX VS 43.1Hz on PC Single 2080Ti) is measured along with at most 1\% mAP drop. Moreover, the jetson results are evidence that our proposed method is feasible for edge orchestration due to the consistent, continuous trade-off between time efficiency and model accuracy with low energy consumption. Fig.\ref{fig:deploy} shows the qualitative example of on-infrastructure detection from TensorRT-format Harmonic PointPillar on Jetson Xavier TX.}  

\section{Conclusion}\label{sec5}
\mm{In this paper, an inconsistency problem is formulated consistently to meet the comprehensive results compared to the state-of-the-art methods. Simulation outcomes show that the proposed method achieved better results on-edge 3D object detection for strengthening the V2X frameworks. First, the foremost cause of the inconsistency among classification, localization and direction estimation is analyzed and derived theoretically and mathematically. Second, the proposed 3D harmonic loss function effectively resolved the inconsistency problem in the pointcloud domain and achieved higher mAP with 75.4 Hz deployment speed than the baseline models. Mathematical derivatives are amended to claim the effectiveness of our proposed loss mechanism. Comprehensive experiment results demonstrate that our proposed method resolved the inconsistency problem to a large extent and achieved detection accuracy on average with no extra inference time cost. Our future work will focus on improving on-infrastructure detection.
}

\section*{Acknowledgments}
\justifying 
This work was supported in part of Basic Science Research Programs of the Ministry of Education (NRF-2018R1A2B6005105) and in part by the National Research Foundation of Korea (NRF) grant funded by the Korea government (MSIT) (No.2019R1A5A8080290).
\bibliographystyle{IEEEtran}
\bibliography{kgn}

\begin{thebibliography}{10}
\providecommand{\url}[1]{#1}
\csname url@samestyle\endcsname
\providecommand{\newblock}{\relax}
\providecommand{\bibinfo}[2]{#2}
\providecommand{\BIBentrySTDinterwordspacing}{\spaceskip=0pt\relax}
\providecommand{\BIBentryALTinterwordstretchfactor}{4}
\providecommand{\BIBentryALTinterwordspacing}{\spaceskip=\fontdimen2\font plus
\BIBentryALTinterwordstretchfactor\fontdimen3\font minus
  \fontdimen4\font\relax}
\providecommand{\BIBforeignlanguage}[2]{{%
\expandafter\ifx\csname l@#1\endcsname\relax
\typeout{** WARNING: IEEEtran.bst: No hyphenation pattern has been}%
\typeout{** loaded for the language `#1'. Using the pattern for}%
\typeout{** the default language instead.}%
\else
\language=\csname l@#1\endcsname
\fi
#2}}
\providecommand{\BIBdecl}{\relax}
\BIBdecl

\bibitem{g3}
C.~E. Palazzi, M.~Roccetti, and S.~Ferretti, ``An intervehicular communication
  architecture for safety and entertainment,'' \emph{IEEE Transactions on
  Intelligent Transportation Systems}, vol.~11, no.~1, pp. 90--99, 2009.

\bibitem{gfl}
X.~Li, W.~Wang, and Wu, ``Generalized focal loss: Learning qualified and
  distributed bounding boxes for dense object detection,'' \emph{Advances in
  Neural Information Processing Systems}, vol.~33, pp. 21\,002--21\,012, 2020.

\bibitem{gflv2}
X.~Li and W.~Wang, ``Generalized focal loss v2: Learning reliable localization
  quality estimation for dense object detection,'' in \emph{Proceedings of the
  IEEE/CVF Conference on Computer Vision and Pattern Recognition}, 2021, pp.
  11\,632--11\,641.

\bibitem{harmony}
K.~Wang and L.~Zhang, ``Reconcile prediction consistency for balanced object
  detection,'' in \emph{Proceedings of the IEEE/CVF International Conference on
  Computer Vision}, 2021, pp. 3631--3640.

\bibitem{paa}
K.~Kim and H.~S. Lee, ``Probabilistic anchor assignment with iou prediction for
  object detection,'' in \emph{European Conference on Computer Vision}.\hskip
  1em plus 0.5em minus 0.4em\relax Springer, 2020, pp. 355--371.

\bibitem{second}
Y.~Yan, Y.~Mao, and B.~Li, ``Second: Sparsely embedded convolutional
  detection,'' \emph{Sensors}, vol.~18, no.~10, p. 3337, 2018.

\bibitem{pointgnn}
W.~Shi and R.~Rajkumar, ``Point-gnn: Graph neural network for 3d object
  detection in a point cloud,'' in \emph{Proceedings of the IEEE/CVF conference
  on computer vision and pattern recognition}, 2020, pp. 1711--1719.

\bibitem{pointrcnn}
S.~Shi, X.~Wang, and H.~Li, ``Pointrcnn: 3d object proposal generation and
  detection from point cloud,'' in \emph{Proceedings of the IEEE/CVF conference
  on computer vision and pattern recognition}, 2019, pp. 770--779.

\bibitem{3dssd}
Z.~Yang, Y.~Sun, S.~Liu, and J.~Jia, ``3dssd: Point-based 3d single stage
  object detector,'' in \emph{Proceedings of the IEEE/CVF conference on
  computer vision and pattern recognition}, 2020, pp. 11\,040--11\,048.

\bibitem{voxelnet}
Y.~Zhou and O.~Tuzel, ``Voxelnet: End-to-end learning for point cloud based 3d
  object detection,'' in \emph{Proceedings of the IEEE conference on computer
  vision and pattern recognition}, 2018, pp. 4490--4499.

\bibitem{voxelrcnn}
J.~Deng, S.~Shi, P.~Li, W.~Zhou, Y.~Zhang, and H.~Li, ``Voxel r-cnn: Towards
  high performance voxel-based 3d object detection,'' in \emph{Proceedings of
  the AAAI Conference on Artificial Intelligence}, vol.~35, no.~2, 2021, pp.
  1201--1209.

\bibitem{part2}
S.~Shi and Z.~Wang, ``From points to parts: 3d object detection from point
  cloud with part-aware and part-aggregation network,'' \emph{IEEE Transactions
  on Pattern Analysis and Machine Intelligence}, vol.~43, no.~8, pp.
  2647--2664, 2021.

\bibitem{pvrcnn}
S.~Shi and C.~Guo, ``Pv-rcnn: Point-voxel feature set abstraction for 3d object
  detection,'' in \emph{Proceedings of the IEEE/CVF Conference on Computer
  Vision and Pattern Recognition}, 2020, pp. 10\,529--10\,538.

\bibitem{voxel-set}
C.~He, R.~Li, S.~Li, and L.~Zhang, ``Voxel set transformer: A set-to-set
  approach to 3d object detection from point clouds,'' in \emph{Proceedings of
  the IEEE/CVF Conference on Computer Vision and Pattern Recognition}, 2022,
  pp. 8417--8427.

\bibitem{sessd}
W.~Zheng, W.~Tang, L.~Jiang, and C.-W. Fu, ``Se-ssd: Self-ensembling
  single-stage object detector from point cloud,'' in \emph{Proceedings of the
  IEEE/CVF Conference on Computer Vision and Pattern Recognition}, 2021, pp.
  14\,494--14\,503.

\bibitem{density-voxel}
J.~S. Hu, T.~Kuai, and S.~L. Waslander, ``Point density-aware voxels for lidar
  3d object detection,'' in \emph{Proceedings of the IEEE/CVF Conference on
  Computer Vision and Pattern Recognition}, 2022, pp. 8469--8478.

\bibitem{centerpoint}
T.~Yin, X.~Zhou, and P.~Krahenbuhl, ``Center-based 3d object detection and
  tracking,'' in \emph{Proceedings of the IEEE/CVF conference on computer
  vision and pattern recognition}, 2021, pp. 11\,784--11\,793.

\bibitem{ciassd}
W.~Zheng, W.~Tang, S.~Chen, L.~Jiang, and C.-W. Fu, ``Cia-ssd: Confident
  iou-aware single-stage object detector from point cloud,'' in
  \emph{Proceedings of the AAAI conference on artificial intelligence},
  vol.~35, no.~4, 2021, pp. 3555--3562.

\bibitem{cl3d}
C.~Lin and D.~Tian, ``Cl3d: Camera-lidar 3d object detection with point feature
  enhancement and point-guided fusion,'' \emph{IEEE Transactions on Intelligent
  Transportation Systems}, 2022.

\bibitem{pp}
A.~H. Lang and S.~Vora, ``Pointpillars: Fast encoders for object detection from
  point clouds,'' in \emph{Proceedings of the IEEE/CVF Conference on Computer
  Vision and Pattern Recognition}, 2019, pp. 12\,697--12\,705.

\bibitem{pixor}
B.~Yang, W.~Luo, and R.~Urtasun, ``Pixor: Real-time 3d object detection from
  point clouds,'' in \emph{Proceedings of the IEEE conference on Computer
  Vision and Pattern Recognition}, 2018, pp. 7652--7660.

\bibitem{bevdet}
S.~Mohapatra and S.~Yogamani, ``Bevdetnet: bird's eye view lidar point cloud
  based real-time 3d object detection for autonomous driving,'' in \emph{2021
  IEEE International Intelligent Transportation Systems Conference
  (ITSC)}.\hskip 1em plus 0.5em minus 0.4em\relax IEEE, 2021, pp. 2809--2815.

\bibitem{fast}
Y.~Chen, S.~Liu, X.~Shen, and J.~Jia, ``Fast point r-cnn,'' in
  \emph{Proceedings of the IEEE/CVF International Conference on Computer
  Vision}, 2019, pp. 9775--9784.

\bibitem{rt3d}
Y.~Zeng, Y.~Hu, and S.~Liu, ``Rt3d: Real-time 3-d vehicle detection in lidar
  point cloud for autonomous driving,'' \emph{IEEE Robotics and Automation
  Letters}, vol.~3, no.~4, pp. 3434--3440, 2018.

\bibitem{ppdeploy}
L.~St{\"a}cker and J.~Fei, ``Deployment of deep neural networks for object
  detection on edge ai devices with runtime optimization,'' in
  \emph{Proceedings of the IEEE/CVF International Conference on Computer
  Vision}, 2021, pp. 1015--1022.

\bibitem{kt}
A.~Geiger, P.~Lenz, and R.~Urtasun, ``Are we ready for autonomous driving? the
  kitti vision benchmark suite,'' in \emph{2012 IEEE conference on computer
  vision and pattern recognition}.\hskip 1em plus 0.5em minus 0.4em\relax IEEE,
  2012, pp. 3354--3361.

\bibitem{dair}
H.~Yu and Luo, ``Dair-v2x: A large-scale dataset for vehicle-infrastructure
  cooperative 3d object detection,'' in \emph{Proceedings of the IEEE/CVF
  Conference on Computer Vision and Pattern Recognition}, 2022, pp.
  21\,361--21\,370.

\bibitem{pointnet}
C.~R. Qi, H.~Su, K.~Mo, and L.~J. Guibas, ``Pointnet: Deep learning on point
  sets for 3d classification and segmentation,'' in \emph{Proceedings of the
  IEEE conference on computer vision and pattern recognition}, 2017, pp.
  652--660.

\bibitem{trs}
J.~Mao, Y.~Xue, and M.~Niu, ``Voxel transformer for 3d object detection,'' in
  \emph{Proceedings of the IEEE/CVF International Conference on Computer
  Vision}, 2021, pp. 3164--3173.

\bibitem{attention}
A.~Vaswani, N.~Shazeer, and N.~Parmar, ``Attention is all you need,''
  \emph{Advances in neural information processing systems}, vol.~30, 2017.

\bibitem{fpointnet}
C.~R. Qi, W.~Liu, C.~Wu, H.~Su, and L.~J. Guibas, ``Frustum pointnets for 3d
  object detection from rgb-d data,'' in \emph{Proceedings of the IEEE
  conference on computer vision and pattern recognition}, 2018, pp. 918--927.

\bibitem{ff}
H.~Zhang, D.~Yang, E.~Yurtsever, K.~A. Redmill, and {\"U}.~{\"O}zg{\"u}ner,
  ``Faraway-frustum: Dealing with lidar sparsity for 3d object detection using
  fusion,'' in \emph{2021 IEEE International Intelligent Transportation Systems
  Conference (ITSC)}.\hskip 1em plus 0.5em minus 0.4em\relax IEEE, 2021, pp.
  2646--2652.

\bibitem{clocs}
S.~Pang, D.~Morris, and H.~Radha, ``Clocs: Camera-lidar object candidates
  fusion for 3d object detection,'' in \emph{2020 IEEE/RSJ International
  Conference on Intelligent Robots and Systems (IROS)}.\hskip 1em plus 0.5em
  minus 0.4em\relax IEEE, 2020, pp. 10\,386--10\,393.

\bibitem{IF}
P.~An, J.~Liang, K.~Yu, B.~Fang, and J.~Ma, ``Deep structural information
  fusion for 3d object detection on lidar--camera system,'' \emph{Computer
  Vision and Image Understanding}, vol. 214, p. 103295, 2022.

\bibitem{ppint}
S.~Vora and A.~H. Lang, ``Pointpainting: Sequential fusion for 3d object
  detection,'' in \emph{Proceedings of the IEEE/CVF conference on computer
  vision and pattern recognition}, 2020, pp. 4604--4612.

\bibitem{focal-loss}
T.-Y. Lin, P.~Goyal, R.~Girshick, K.~He, and P.~Doll{\'a}r, ``Focal loss for
  dense object detection,'' in \emph{Proceedings of the IEEE international
  conference on computer vision}, 2017, pp. 2980--2988.

\bibitem{focalloss}
------, ``Focal loss for dense object detection,'' in \emph{Proceedings of the
  IEEE international conference on computer vision}, 2017, pp. 2980--2988.

\bibitem{fasterrcnn}
S.~Ren, K.~He, R.~Girshick, and J.~Sun, ``Faster r-cnn: Towards real-time
  object detection with region proposal networks,'' \emph{Advances in neural
  information processing systems}, vol.~28, 2015.

\bibitem{mmdet3d}
M.~Contributors, ``{MMDetection3D: OpenMMLab} next-generation platform for
  general {3D} object detection,''
  \url{https://github.com/open-mmlab/mmdetection3d}, 2020.

\bibitem{pircnn}
L.~Xie, C.~Xiang, and Yu, ``Pi-rcnn: An efficient multi-sensor 3d object
  detector with point-based attentive cont-conv fusion module,'' in
  \emph{Proceedings of the AAAI conference on artificial intelligence},
  vol.~34, no.~07, 2020, pp. 12\,460--12\,467.

\bibitem{eqpvrcnn}
Z.~Yang, L.~Jiang, Y.~Sun, B.~Schiele, and J.~Jia, ``A unified query-based
  paradigm for point cloud understanding,'' in \emph{Proceedings of the
  IEEE/CVF Conference on Computer Vision and Pattern Recognition}, 2022, pp.
  8541--8551.

\bibitem{tanet}
Z.~Liu, X.~Zhao, T.~Huang, R.~Hu, Y.~Zhou, and X.~Bai, ``Tanet: Robust 3d
  object detection from point clouds with triple attention,'' in
  \emph{Proceedings of the AAAI Conference on Artificial Intelligence},
  vol.~34, no.~07, 2020, pp. 11\,677--11\,684.

\end{thebibliography}
\end{document}